\documentclass{article}
\usepackage[preprint]{neurips_2021}
\usepackage[utf8]{inputenc} 
\usepackage[T1]{fontenc}    
\usepackage{hyperref}       
\usepackage{url}            
\usepackage{booktabs}       
\usepackage{amsfonts}       
\usepackage{nicefrac}       
\usepackage{microtype}      
\usepackage{xcolor}         
\usepackage{todonotes}
\usepackage{amsmath, amsthm}
\usepackage{booktabs}

\usepackage{listings}
\definecolor{codegreen}{rgb}{0,0.6,0}
\definecolor{codegray}{rgb}{0.5,0.5,0.5}
\definecolor{codepurple}{rgb}{0.58,0,0.82}
\definecolor{backcolour}{rgb}{0.95,0.95,0.92}

\lstdefinestyle{mystyle}{
    backgroundcolor=\color{backcolour},   
    commentstyle=\color{codegreen},
    keywordstyle=\color{magenta},
    numberstyle=\tiny\color{codegray},
    stringstyle=\color{codepurple},
    basicstyle=\ttfamily\footnotesize,
    breakatwhitespace=false,         
    breaklines=true,                 
    captionpos=b,                    
    keepspaces=true,                 
    showspaces=false,                
    showstringspaces=false,
    showtabs=false,                  
    tabsize=2
}

\usepackage{ulem}

\def\linf{L_{\infty}}

\title{Scaling Laws for Neural Machine Translation}

\author{%
Behrooz Ghorbani\\\texttt{ghorbani@google.com}\And
Orhan Firat\\\texttt{orhanf@google.com}\AND
Markus Freitag\\\texttt{freitag@google.com}\And
Ankur Bapna\\\texttt{ankurbpn@google.com}\And
Maxim Krikun\\\texttt{krikun@google.com}\And
Xavier Garcia\\\texttt{xgarcia@google.com}\And
Ciprian Chelba\\\texttt{ciprianchelba@google.com}\And
Colin Cherry\\\texttt{colincherry@google.com} \\~
}

\usepackage{todonotes}

\newtheorem{prop}{Proposition}

\begin{document}

\maketitle

\begin{abstract}
We present an empirical study of scaling properties of encoder-decoder Transformer models used in neural machine translation (NMT).
We show that cross-entropy loss as a function of model size follows a certain scaling law. Specifically 
(i) We propose a formula which describes the scaling behavior of cross-entropy loss as a bivariate function of encoder and decoder size, and show that it gives accurate predictions under a variety of scaling approaches and languages; we show that the total number of parameters alone is not sufficient for such purposes.
(ii) We observe different power law exponents when scaling the decoder vs scaling the encoder, and provide recommendations for optimal allocation of encoder/decoder capacity based on this observation. (iii) We also report that the scaling behavior of the model is acutely influenced by \textit{composition bias} of the train/test sets, which we define as any deviation from naturally generated text (either via machine generated or human translated text). We observe that natural text on the target side enjoys scaling, which manifests as  successful reduction of the cross-entropy loss. 
(iv) Finally, we investigate the relationship between the cross-entropy loss and the quality of the generated translations. We find two different behaviors, depending on the nature of the test data. For test sets which were originally translated from target language to source language, both loss and BLEU score improve as model size increases. In contrast, for test sets originally translated from source language to target language, the loss improves, but the BLEU score stops improving after a certain threshold.
We release generated text from all models used in this study. 
\end{abstract}

\section{Introduction} 

Scaling properties of neural networks have long been an intriguing topic of study \citep{10.5555/2969735.2969754,Amari1992FourTO}. Along with the practical success of modern neural networks at scale, theoretical understanding of the factors governing the quality and training dynamics of large neural networks has also being developing \citep{
advani2017highdimensional,rosenfeld2019constructive,Geiger_2020,ghorbani2020linearized,chobernoulli2020,DBLP:journals/corr/abs-2102-04074,bahri2021explaining,loureiro2021capturing}. In particular, scaling model sizes, datasets and the total computation budget has been identified as a reliable approach to improve generalization performance on several machine learning tasks. For many of these tasks the scaling behavior of neural networks is highly predictable; model fit or test loss can be described precisely as a function of its number of parameters 
\citep{hestness2017deep,kaplan2020scaling,henighan2020scaling,hernandez2021scaling,rosenfeld2019constructive}. Neural machine translation (NMT) has long enjoyed the benefits of scaling \citep{huang2019gpipe,DBLP:journals/corr/abs-1907-05019,DBLP:journals/corr/abs-2006-16668}, but studies investigating the scaling behavior of NMT models are missing. 
We present the first large-scale systematic study of scaling laws for encoder-decoder Transformer models applied to NMT  \citep{DBLP:journals/corr/VaswaniSPUJGKP17}. \footnote{An initial version of this study was submitted to NeurIPS 2021.} \footnote{A few weeks before the publication of this manuscript on Arxiv, \cite{gordondata} appeared on OpenReview. While both papers study scaling laws for NMT, our studies focus on different parameter regimes (393K-56M vs 100M-3.5B).}

We start with highlighting the major differences between decoder-only language models, where the majority of the previous work has focused, and encoder-decoder (conditional) language models applied to NMT. The two differ along a few crucial dimensions. The first difference results from the very nature of the separate architectures being used, i.e. decoder-only vs encoder-decoder. The presence of separate architectural components complicates the study of scaling properties due to the increased degree of freedom.
Second, contrary to language modeling, the task of machine translation is conditional: the task is predictive rather than fully generative.
Furthermore, this prediction task is ambiguous: there is no one right answer for a given source, and translations can vary substantially depending on the translator's incentives.
This manifests itself as different scaling benefits for different test sets. 
To take an extreme example, a test set translated by someone who writes nearly word-for-word translations may benefit less from model scaling than one translated by someone who considers each translation a work of art. 
In this work, these differences in difficulty coincide with the translation direction of the test set; that is,
whether the source was translated into the target (source-original) or vice versa (target-original).
Source-original data has translated text on the target side, which contains several artifacts of ``translationese'' that distinguish it from text originally written in that language, often lacking the diversity and complexity of ``natural'' text \citep{Koppel:2011:TD:2002472.2002636}, while target-original data requires the prediction of more complex natural text on the target side.
Finally, unlike language models, NMT is evaluated on metrics that quantify generation quality against reference translations (for eg. BLEU) \citep{papineni2002bleu} instead of evaluating model fit (perplexity) on an evaluation set. 

In this paper, we aim to provide empirical answers to the following research questions: 

\begin{enumerate}
    \item \textbf{Does the encoder-decoder architecture for NMT share the same scaling law function as the language models?} 
    Contrary to previous work on LM, we show that a univariate law depending on the total number of parameters in the network does not adequately describe the scaling behavior of NMT models. Our scaling laws parameterize the cross entropy loss as a bivariate function of the number of encoder parameters and the number of decoder parameters as separate variables. Our results indicate that the scaling behavior is largely determined by the total capacity of the model, and the capacity allocation between the encoder and the decoder.

    \item \textbf{How does the naturalness of source/target side data affect scaling behavior?} We study the effect of naturalness of the source and target text, both for training and evaluation. 
    When evaluating with target side natural text, scaling the model capacity continues improving model quality throughout our range of measurements. On the other hand, improvements on cross-entropy saturate (or reaches the irreducible error region) on source side natural evaluation sets even for moderately-sized models.
    
    \item \textbf{Do scaling improvements in cross-entropy translate into corresponding improvements in generation quality?} Finally we study the relationship between generation quality and cross-entropy and how their correlation changes as we: (i) Scale different components of the model (encoder vs decoder) and (ii) Evaluate on source-natural or target-natural evaluation sets. 

\end{enumerate}

Our results on multiple language pairs and training/test data compositions validate that \textbf{model scaling predictably improves the cross-entropy on validation data}. However, our findings also raise several questions regarding the effect of naturalness of training and evaluation text and how cross-entropy eventually relates with generation quality for auto-regressive generative models.

\section{Effect of Scaling on Cross-Entropy} \label{sec:exps}
\subsection{Experimental setting}

\paragraph{Model Architectures and Training} 
We train a series of pre-layer norm Transformer networks with varying sizes \cite{xiong2020layer}. Models are trained with per-token cross-entropy loss and Adafactor optimizer \cite{shazeer2018adafactor}. All models are trained with a fixed batch-size of $500$k tokens and dropout rate of $0.1$ for residuals, feed-forward activations and attention. All models are trained to near convergence for $500$k training steps. Details of the model hyper-parameters are described in Appendix \ref{app:hyps}.

\paragraph{Model Scaling} Transformer architecture consists of Transformer Blocks: a cascade of self-attention, cross-attention and feed-forward layers, each having multiple adjustable hyper-parameters (e.g. model-dimension, number of attention heads, attention projection dimension etc.). Considering the combinatorial expansion of the search space for scaling each one \cite{JMLR:v21:20-074,DBLP:conf/nips/LevineWSBS20,wies2021transformer}, in this study we choose to vary only the total number of Transformer Blocks, while keeping the internal hyper-parameters intact across different scales. In other words, we scale the depth of the Transformers while keeping width and other variables fixed. We use GPipe pipeline parallelism for scaling \cite{huang2019gpipe} thanks to its flexible API across various depths.

In an encoder-decoder Transformer architecture for NMT, depth scaling can naturally be implemented by varying encoder-decoder blocks independently or symmetrically. Hence, we examine the change in the cross-entropy loss as the number of parameters increase with three depth scaling approaches:

\begin{description}
\item[] \textit{Encoder Scaling}: vary encoder depth ($2$ to $64$) while decoder depth is fixed ($6$ layers).
\item[] \textit{Decoder Scaling}: vary decoder depth ($2$ to $64$) while encoder depth is fixed ($6$ layers).
\item[] \textit{Symmetric Scaling}: increasing decoder and encoder layers together (from $2$ to $64$), i.e. the number of Transformer Blocks in the encoder and decoder being equal.  
\end{description}
For all experiments, configuration of the individual layers is unchanged: the model dimension, width of the feed-forward layer, and number of attention heads are fixed respectively at $1024$, $8192$, and $16$. \footnote{A complete description of the model architecture is provided in Appendix \ref{app:hyps}} Each encoder layer adds approximately 20M parameters to the model while each decoder layer adds around 25M parameters. In this section, we train $95$ such models which scale the encoder / decoder size by approximately a factor of $32$ (from roughly $40$M parameters to $1.5$B parameters). Following the convention in this literature, we do not count the parameters in the embedding and softmax layers towards the model size.

\paragraph{Language Pairs} We report results on two language pairs, English$\to$German and German$\to$English, using an in-house web-crawled dataset with around $2.2$ billion sentence pairs (approximately 55 billion tokens) for both translation directions. This dataset provides a large enough training set to ensure the dataset size is not a bottleneck in the model performance. 

\paragraph{Evaluation Sets} We use a variety of test sets for evaluation covering different domains: (i) Web-Domain (ii) News-Domain (iii) Wikipedia (iv) Patents.
The news-domain test sets come from the WMT2019~\cite{barrault-EtAl:2019:WMT} evaluation campaign (newstest2019) for all language pairs. The other test sets are internal test sets representing the different domains, ranging from 500 to 5000 sentence pairs. For each domain, we randomly sample sentences in the source language and use professional translators to generate a reference translation in the target language. Throughout the paper, we will refer this type of test sets as \textit{source-original} as the source sentences have been crawled from the web while the reference translations are added later. For most of the domains, we also have a \textit{target-original} counterpart which is generated in the opposite direction: Sentences are crawled in the target language and human translated into the source language. Earlier work \cite{Freitag19,freitag-etal-2020-bleu,graham2020statistical} showed that it is important to differentiate between the two different kinds of test sets as the style of natural sentences and human (or machine) translations (translationese) is quite different.
Cross-entropy loss is evaluated on the different test sets during training. To reduce the variation caused by the parameter fluctuations at the end of the training, we present the median loss over the last $50$k steps of the training as the final loss.

\subsection{Results}
\begin{figure}[h]
  \centering
  \includegraphics[width=\textwidth]{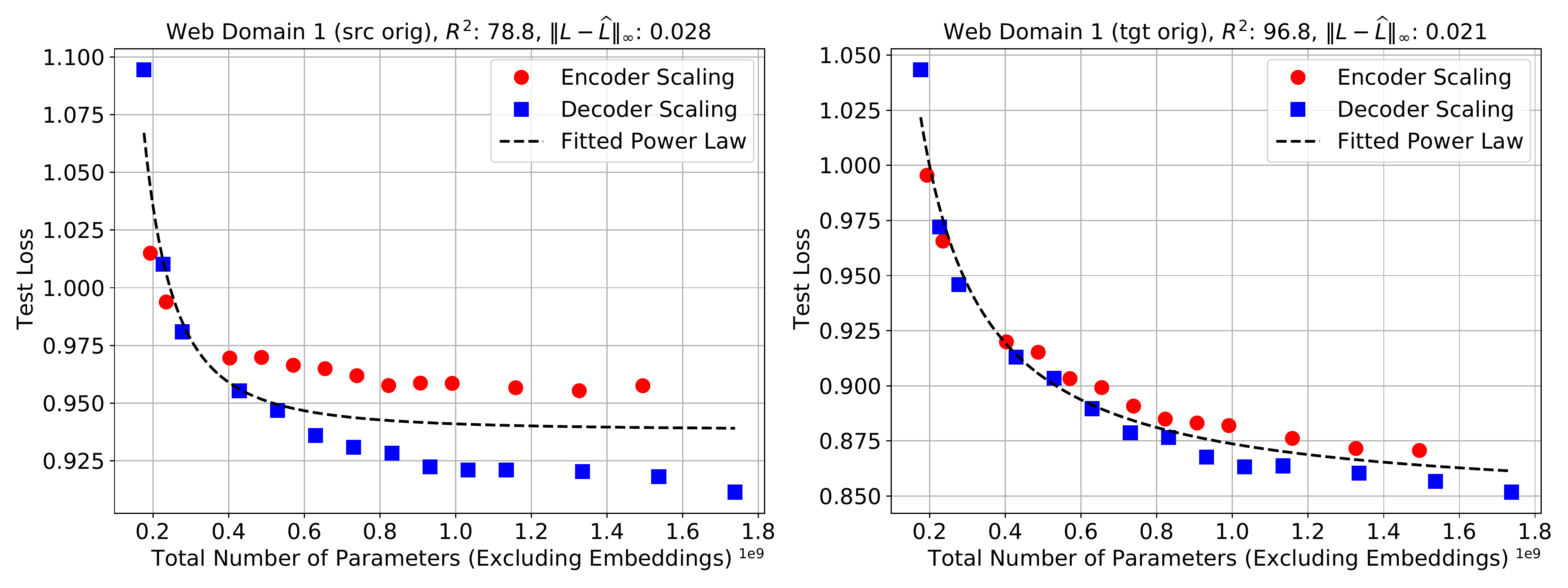}
  \caption{Evolution of the test loss as a function of the total model parameters for English$\to$German. Scaling the encoder has different effects compared to scaling the decoder. As such, traditional power laws of type expressed as in Eq.~\eqref{eq:simple_power_law} are unable to capture the correct scaling behavior. R-squared ($100 \times \frac{\mbox{explained variance}}{\mbox{total variance}}$) and maximum absolute deviation ($\Vert \cdot \Vert_{\infty}$) are reported for each fit. \label{fig:old_scaling}}
\end{figure}

Figure \ref{fig:old_scaling} shows the empirical evolution of the test loss on the Web-Domain test sets for encoder and decoder scaling for English$\to$German. To compare the empirical results with the scaling laws present in the literature for decoder only models \cite{kaplan2020scaling, henighan2020scaling}, we have fitted a power law of the form
\begin{equation} \label{eq:simple_power_law}
    \hat{L}(N) = \alpha N^{-p} + \linf
\end{equation}
to the data. \footnote{Details of the curve fitting procedure are presented in Appendix \ref{app:curve_fitting}.} Here, $N$ is the total number of parameters outside of embedding / softmax layers and $\{\alpha, p, \linf \}$ are the fitted parameters of the power law. As Figure \ref{fig:old_scaling} suggests, scaling the encoder has different effects on the test loss compared to scaling the decoder. As such, simple power-law relations similar to Eq.~\eqref{eq:simple_power_law} that only consider the total number of parameters, fail to capture the correct scaling behavior of the model. 

\paragraph{Proposed Scaling Law} To tackle this issue, we present a new scaling law that reflects the encoder-decoder nature of the architecture as well as the bilingual format of the data. Let $N_e$ and $N_d$ be the number of non-embedding parameters in the encoder and the decoder respectively. Then, our proposed scaling law has the form 
\begin{equation} \label{eq:scaling_law}
    \hat{L}(N_e, N_d) = \alpha \bigg( \frac{\bar{N}_e}{N_e} \bigg)^{p_e} \bigg(\frac{\bar{N}_d}{N_d} \bigg)^{p_d} + L_{\infty}
\end{equation}
where $\{\alpha, p_e, p_d, \linf\}$ are test set specific (fitted) parameters. $\bar{N}_e$ and $\bar{N}_d$ are fixed normalization parameters corresponding to the number of encoder / decoder parameters in our baseline $12$-layer encoder-decoder model.\footnote{Corresponds to $6$-layer encoder - $6$-layer decoder.}  In this formulation, $\alpha$ corresponds to the maximum loss reduction (as compared to the baseline model) that one can hope from scaling, while $p_e$ and $p_d$ are the scaling exponents for encoder and decoder respectively. $\linf$ corresponds to the irreducible loss of the data.

\begin{figure}[t!]
  \centering
  \includegraphics[width=\textwidth]{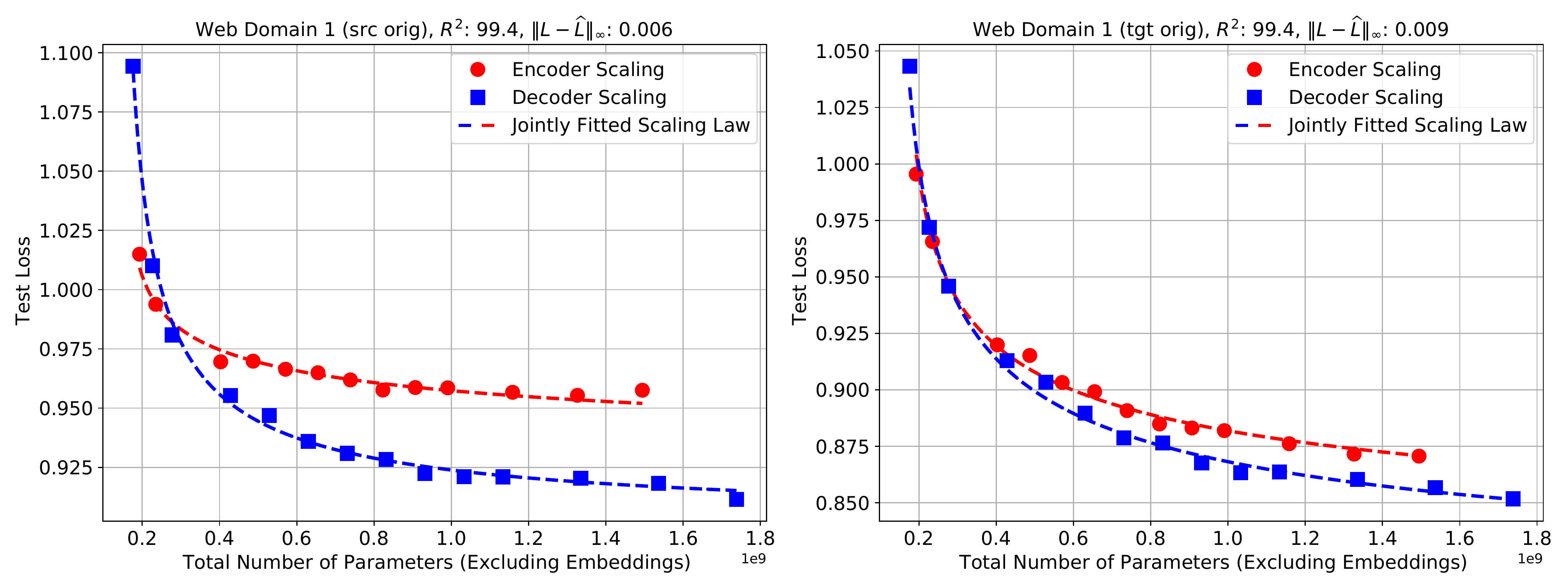}
  \caption{Evolution of log-perplexity as a function of the model size for English$\to$German models. Eq. \eqref{eq:scaling_law} is jointly fitted to the empirical loss values from encoder scaling and decoder scaling experiments. Our proposed scaling law is able to capture more than $99\%$ of the variation in the data. We anticipate some fluctuations around the predicted trend (with estimated standard deviation of $0.003$) caused by the randomness in the training pipeline (see Appendix \ref{app:variation}). We observe similar results for our other test sets (see Figures \ref{fig:EnDe_all} \& \ref{fig:EnDe_all_loglog} of the appendix). \label{fig:scaling_laws_encoder_decoder}} 
\end{figure}

Figure \ref{fig:scaling_laws_encoder_decoder} presents the fit achieved by the proposed scaling law on Web-Domain test sets. The dashed lines describe the fit of the scaling law given in Eq.~\eqref{eq:scaling_law} to the empirical (encoder \& decoder scaling) data. The plots suggest that our proposed scaling law is able to simultaneously capture both encoder and decoder scaling behaviors.

\begin{figure}[ht]
  \centering
  \includegraphics[width=\textwidth]{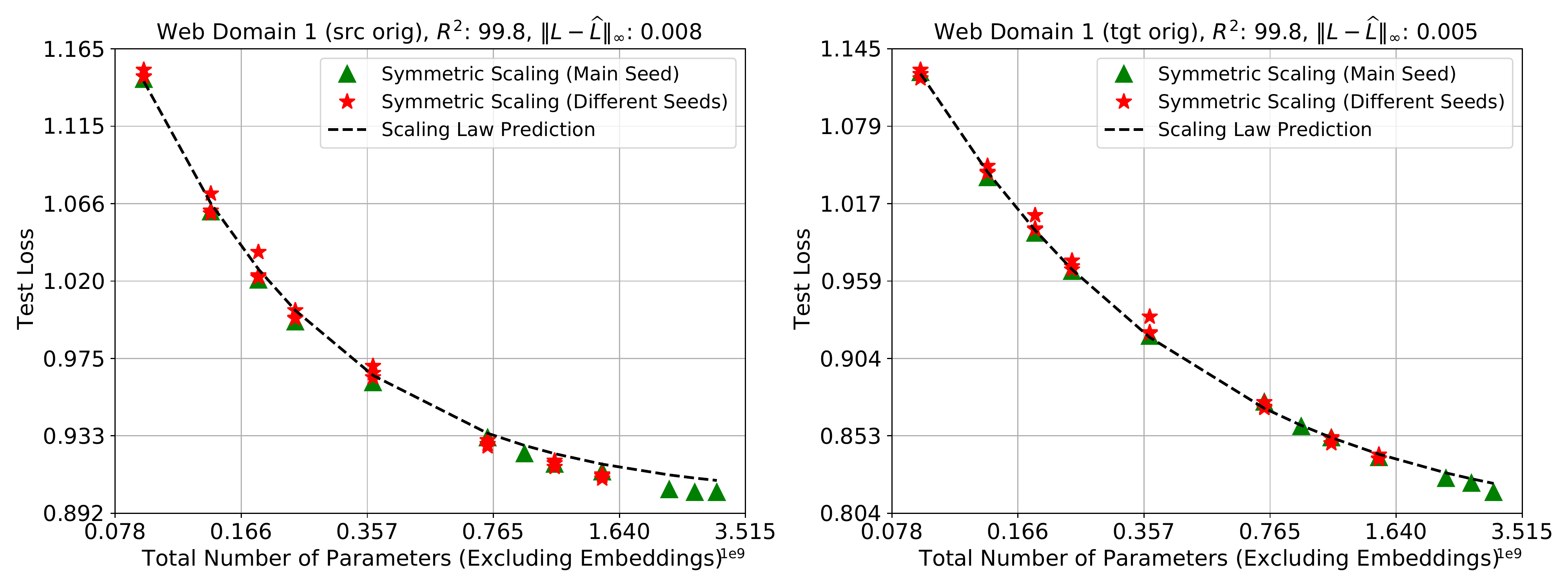}
    \vspace{-1em}
  \caption{Comparison of the (out-of-sample) predictions of the scaling law with the empirical test loss values from symmetric scaling English$\to$German models. Eq. \eqref{eq:scaling_law} is fitted only using the encoder / decoder scaling data and then just evaluated on the symmetric scaling model parameters. Our proposed scaling law is able to almost fully capture the variation in the data ($R^2 = 99.8\%$) even though it has not been fitted on it. To examine the randomness in the results, we have repeated a subset of training runs with $4$ different random seeds (see Appendix \ref{app:variation} for more details). We observe similar results for our other test sets (see Figure \ref{fig:EnDe_all_OOS} of the appendix). \label{fig:scaling_laws_OOS}}
\end{figure}

To validate the (out-of-sample) prediction power of these scaling laws, we compare their predictions with empirical loss values achieved by our symmetric scaling models. Figure \ref{fig:scaling_laws_OOS} presents this comparison. The plots suggest that the predictions of the scaling law match the empirical (out-of-sample) results with remarkable accuracy. These results suggest that the predictions of the scaling law are not sensitive to the scaling approach; the scaling law fitted on encoder / decoder scaling data is able to almost perfectly predict the scaling behavior of symmetric models. Notice that the parameter range of symmetric scaling models is much larger than either of the encoder or decoder scaling models. Nevertheless, our fitted scaling laws are able to extrapolate effectively to models with sizes beyond the ones used for fitting them.

\begin{figure}[h]
  \centering
  \includegraphics[width=\textwidth]{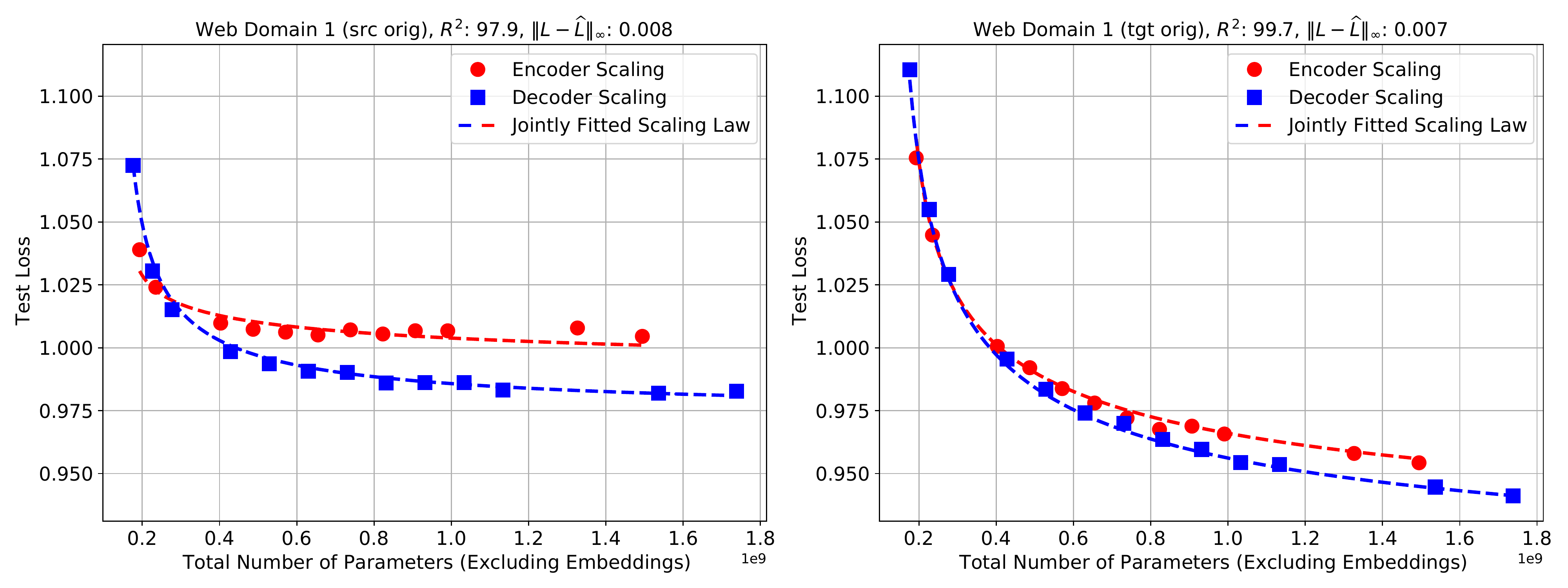}
    \vspace{-1em}
  \caption{Fitted scaling laws for De$\rightarrow$En translation task. The scaling law \eqref{eq:scaling_law} is jointly fitted to the empirical loss value from encoder scaling and decoder scaling experiments. Similar to En$\to$De case, the law is able to describe the empirical scaling behavior of the models with high accuracy. See Figures \ref{fig:DeEn_all} \& \ref{fig:DeEn_all_loglog} in the appendix for the fit on other test sets. \label{fig:DeEn_snapshot}}
\end{figure}

Figures \ref{fig:scaling_laws_encoder_decoder} \& \ref{fig:scaling_laws_OOS} suggest that the functional form proposed in Eq.~\eqref{eq:scaling_law} captures the scaling behavior of English$\rightarrow$German models accurately. To verify the robustness of our proposed scaling law, we evaluate it on an additional translation task namely, German$\to$English (De$\rightarrow$En). 
Figure~\ref{fig:DeEn_snapshot} depicts the corresponding scaling behavior on Web-Domain test set. Similar to the En$\rightarrow$De case, our proposed functional form is able to closely capture the scaling behavior of the models. 

\subsection{Analysis} 
\label{subsec:analysis}

\begin{figure}[h]
  \centering
  \includegraphics[width=\textwidth]{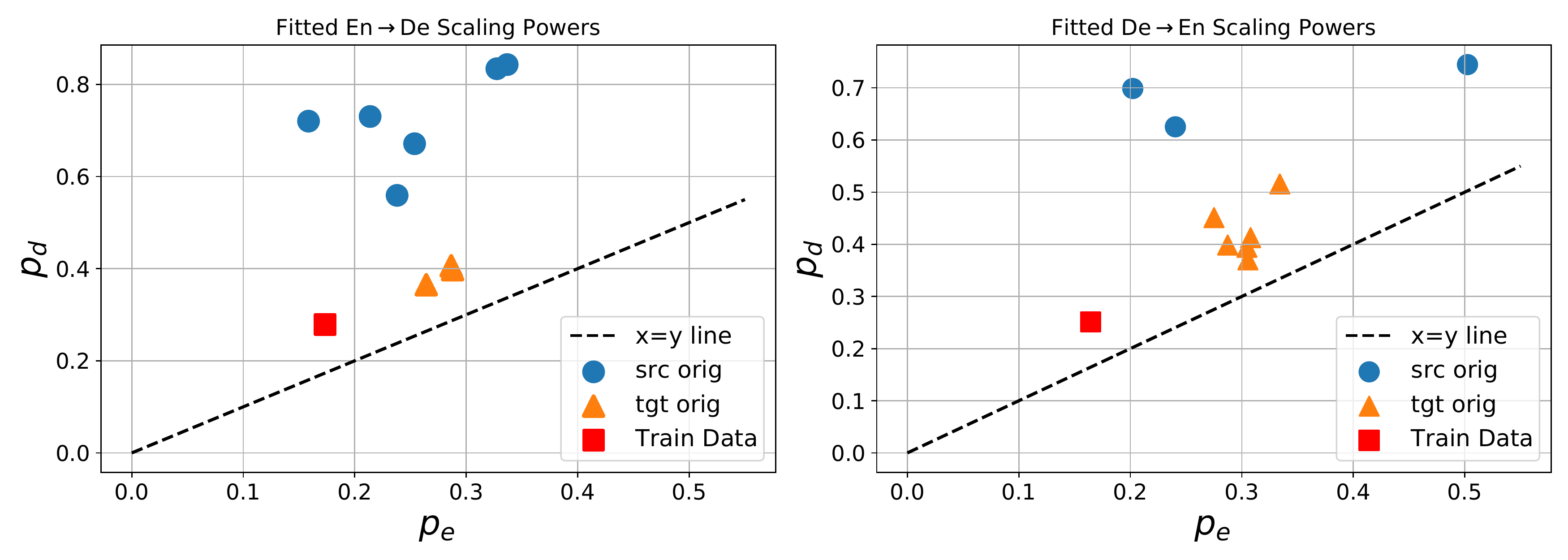}
    \vspace{-1em}
  \caption{Fitted scaling law exponents for all our test sets. Across all the test sets under consideration, we observe $p_d > p_e$. \label{fig:EnDe_coeff}}
\end{figure}

The above results suggest that scaling law formalized in Eq.~\eqref{eq:scaling_law} captures the scaling behavior of the Transformer NMT models in multiple language pairs. As such, we can study the fitted coefficients to fully understand the scaling properties of these models. Figure \ref{fig:EnDe_coeff} presents the fitted coefficients for all of the test sets under consideration. Several observations are in order:
\paragraph{Decoder vs Encoder Scaling:} On all our test sets, the decoder exponents were observed to be larger than the encoder exponents, $p_d > p_e$. As a result, when improving the test loss is concerned, it is much more effective to scale the decoder rather than the encoder. This is contrary to the usual practice; due to latency considerations, many practitioners train NMT models with deep encoders and shallow decoders \cite{kasai2021deep}. Our results suggest this practice could be sub-optimal in terms of loss reduction. Proposition \ref{prop:optimal_scaling} below leverages Eq.~\eqref{eq:scaling_law} to provide guidance on how to allocate parameters in between the encoder and decoder optimally. The proof is presented in Appendix \ref{app:proofs}.

\begin{prop}[Optimal Scaling] \label{prop:optimal_scaling}
Assume the loss performance of the model is described by Eq. \eqref{eq:scaling_law}. Let $B$ denote the budget for total number of parameters. Then, the optimal encoder / decoder sizes (denoted respectively by $N_e^*$ and $N_d^*$) are:
\begin{align} \label{eq:optimal_params}
    N_e^* = \frac{p_e}{p_e + p_d} B, \qquad N_d^* = \frac{p_d}{p_e + p_d} B.
\end{align}
In addition, when optimally scaling the model, the scaling law reduces to:
\begin{align}\label{eq:optimal_scaling}
     \hat{L}_{opt}(B) = \alpha^* B^{-(p_d + p_e)} + \linf, \qquad \alpha^* \equiv \alpha \bigg(\frac{\bar{N}_e (p_e + p_d)}{p_e} \bigg)^{p_e} \bigg(\frac{\bar{N}_d (p_e + p_d)}{p_d} \bigg)^{p_d}.
\end{align}
\end{prop}

\begin{figure}[h]
  \centering
  \includegraphics[width=\textwidth]{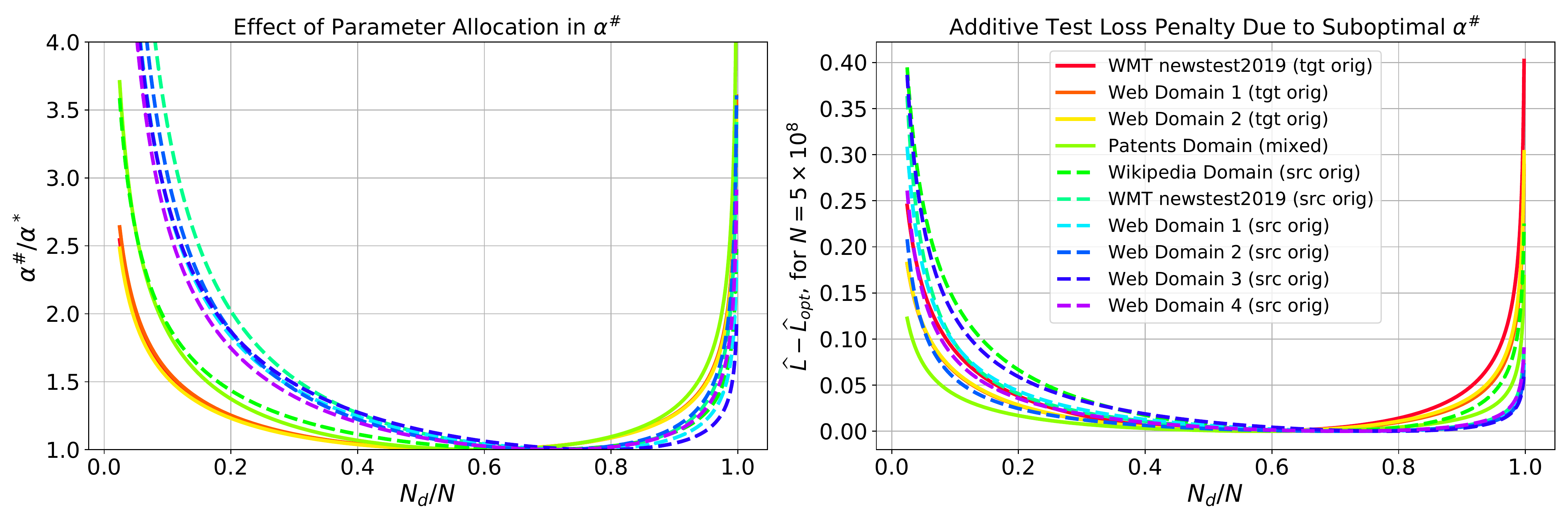}
    \vspace{-1em}
  \caption{We use our fitted scaling laws to evaluate the effect of encoder / decoder parameter allocation ratio when proportionally scaling the encoder and the decoder. Left: $\alpha^\# / \alpha*$ for different parameter allocation schemes. Right: The predicted additive loss penalty, $(\widehat{L} - \widehat{L}_{opt})$, for a model with $5\times 10^8$ total (non-embedding) parameters. Each line corresponds to a different test set. \label{fig:linear_scaling}}
\end{figure}

Proposition \ref{prop:optimal_scaling} suggests that when $\frac{N_e}{N_d} = \frac{p_e}{p_d}$, one can achieve the best possible scaling behavior for the task. Inspection of the functional form of Eq.~\eqref{eq:scaling_law} suggests that as long as $N_d / N_e$ is fixed as the model scales (i.e. the encoder and decoder grow proportionally together), the optimal scaling exponent, $(p_e + p_d)$, can be achieved, albeit with a potentially sub-optimal multiplicative constant, $\alpha^\#$. To examine how significant this sub-optimality can be, in Figure \ref{fig:linear_scaling}, we compare the multiplicative constants resulting from proportional scaling of the encoder and decoder with different values of $N_d / N_e$. The results suggest that as long as the parameter allocation is not extremely far from $(N_e^*, N_d^*)$, the scaling scheme is approximately optimal. In particular, symmetrically scaling the encoder and decoder layers, which yields $N_d / N \approx 0.55$, is barely distinguishable from the optimal scaling scheme described in Proposition \ref{prop:optimal_scaling}. In contrast, lopsided models which heavily favor the encoder or the decoder achieve a much worse multiplicative constant.  

\section{Effect of Dataset Composition Bias on Scaling Behavior}

Translation deals with the problem of mapping a sequence in one language into another language. 
A good translation should not only be adequate and fluent, but should ideally also adopt the style of a sentence naturally written in the target language.
This necessitates MT models to make sense of natural looking inputs and generate natural looking outputs. As mentioned in Section~\ref{sec:exps}, the examples used to train or test NMT models carry a critical bias, which we refer to as \textit{composition bias}. Composition bias is introduced because of the unavailability of source-target examples (pairs) that are both natural in the accessible data generating distribution. For any given naturally generated text in a language, the corresponding text in the other language is either translated by humans, introducing \textit{translationese} bias or translated by other machine translation systems, introducing \textit{MT} bias. We consider both biases affecting the problem from a similar angle, hence we bundle them and call it composition bias. While machine translation by design has composition bias in the training/test sets employed \cite{freitag-etal-2020-bleu,riley2020translationese}, its effect on model scaling is unknown. In this section we investigate the role of composition bias in scaling and identify critical factors playing role.

We caution the reader to not take the composition bias as a problem specific to NMT. In fact as most training corpora in NMT are web-crawled, they can contain machine translation output on either the source or target side. Considering the growth of generated content in the web by machine learning models \footnote{https://openai.com/blog/gpt-3-apps/}~\footnote{https://blog.google/products/translate/one-billion-installs/}, it is not improbable that a proportion of the content collected and used by machine learning models is going to be biased by other models that are continuously generating content.

\paragraph{The Effect of Test Set Construction:} We will first take a look at the impact of composition bias on the test sets used in this study and then investigate the influence on the training set. Figure \ref{fig:EnDe_coeff} shows the fitted scaling law
coefficient for all of our test sets. The coefficients suggests that the scaling powers for source-original test sets are drastically different from those of target-original test sets. This behavior is in direct contrast with language modeling setting \cite{kaplan2020scaling} where it was observed that the evaluation on different test sets merely acted as a scaling penalty that only changed the multiplicative constants of the scaling law.

\begin{figure}[h]
  \centering
  \includegraphics[width=\textwidth]{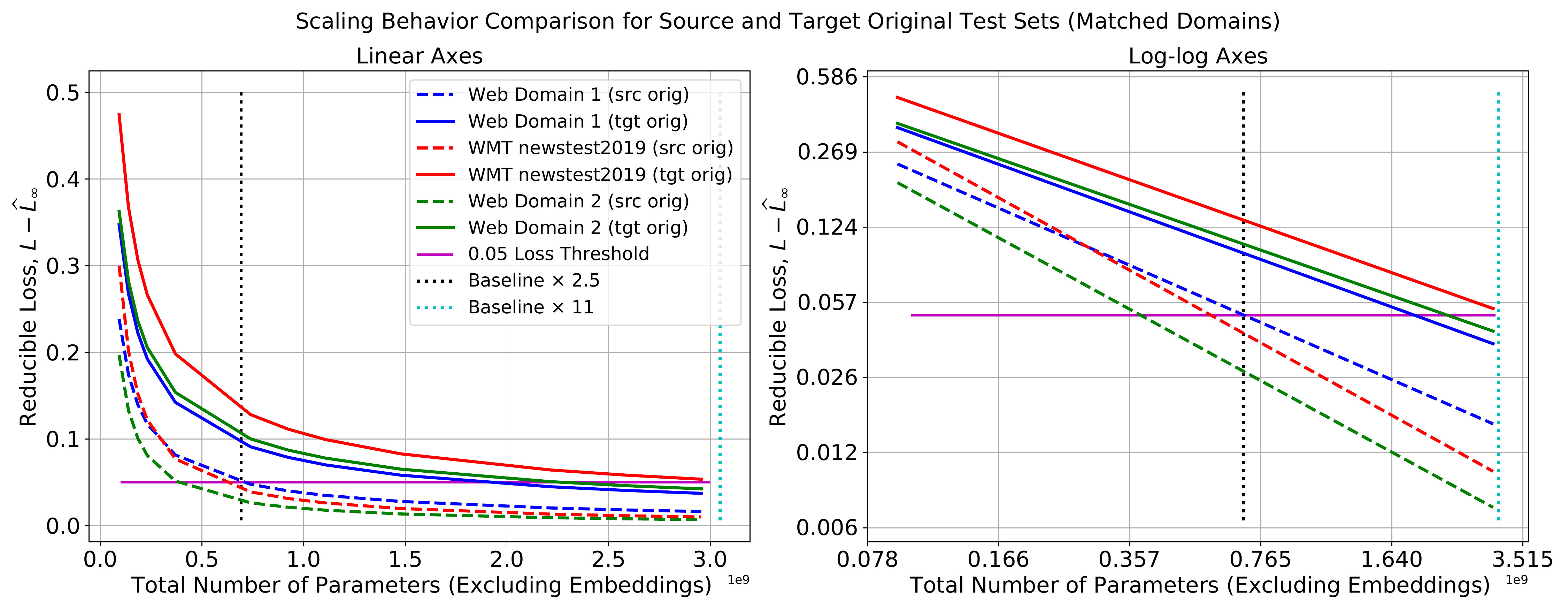}
  \caption{A comparison of scaling behavior across source and target original test sets. We use our fitted scaling laws to estimate the evolution of reducible loss for each test set. All scaling trends correspond to symmetrically scaling the encoder and decoder layers.  \label{fig:EnDe_Test_Comparison}}
\end{figure}

To elucidate this phenomenon further, in Figure \ref{fig:EnDe_Test_Comparison}, we compare the scaling trends for different source and target original test sets. To factor out the effect of the data domain, we present one source original and one target original test set for each domain. Several observations are in order: Test sets with a similar composition approach (source or target original) have a qualitatively similar scaling behavior. However, scaling behavior is vastly different between the two composition approaches. Reducible loss quickly decays to zero for source original test sets. In fact, we observe that scaling our baseline 6L-6L model by a factor of $2.5$ is sufficient for ensuring that reducible loss is below $0.05$ for all source original test sets. In contrast, on target original test sets, the loss decays much more slowly with model size. For comparison, to ensure that reducible loss is below $0.05$ for all target original test sets, we estimate that the baseline model has to be scaled up by a factor of $11$.

Because of this behavior, the value of larger models in NMT is closely tied to their evaluation sets: On source original test sets, due to larger scaling exponents, even moderate increases in model size are sufficient for pushing the reducible loss close to zero. Hence, beyond a few hundred million parameters, there is no benefit in increasing the model size. In contrast, for target original test sets, which generally have smaller scaling exponents, large models are needed to push the reducible loss to zero.

\paragraph{The Effect of Training Set Construction:} The results of the previous section suggest that the construction of the test data plays a key role in the scaling behavior of the model. Now, we briefly examine the role of \emph{training data} construction on the scaling behavior. To do this, we generate two En$\rightarrow$De datasets, that were not used in the previous experiments. One fully target original and another completely source original. 

To generate the target original dataset, we compile a set of German documents from the web. Documents are screened to ensure the data is not machine generated. We use a Hybrid model (with 380M parameters) \cite{chen2018best} to back-translate (BT) these documents to English. Similarly, for the source original data, we collect human generated English documents and (forward) translate them to German using a hybrid model (with approximately 327M parameters). Both datasets provide us with approximately 2.2 billion training examples. We mimic the experimental setup of Section \ref{sec:exps}.

Note that even though these datasets are not human generated, they reflect important aspects of training large NMT models. Many modern NMT datasets are harvested from the web and as a result, are contaminated with machine generated data. Moreover, many popular data augmentation algorithms such as Back Translation \cite{sennrich2016improving}, sequence level distillation \cite{kim2016sequencelevel} and self training \cite{he2020revisiting} purposefully add machine generated data into the training pipeline in order to take advantage of monolingual data.

\begin{figure}[h]
  \centering
  \includegraphics[width=\textwidth]{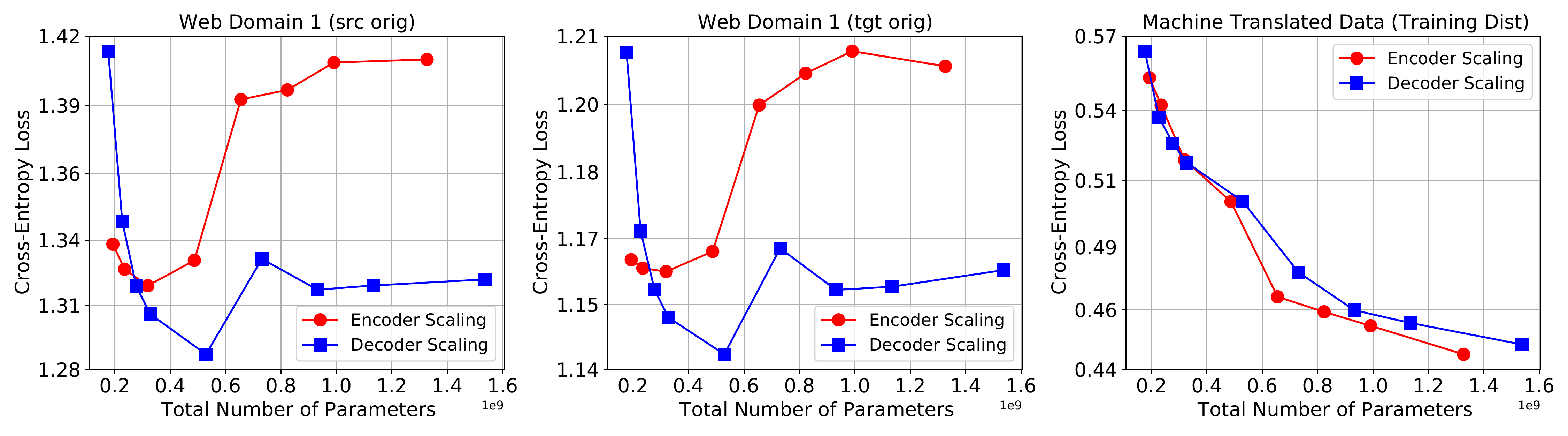}
    \vspace{-1.5em}
  \caption{Scaling behavior of models trained on back-translated data. Right: Increasing the model size successfully reduces the loss on the training distribution. However, on the test data (left and center) the loss increases after approximately $400M$ parameters. \label{fig:BT_data}}
\end{figure}

Figure \ref{fig:BT_data} describes the scaling behavior for models trained on target-original data. We observe that even though larger models are successful in reducing the training loss, they are unable to improve the test loss after roughly $400$M parameters. Once this size threshold is exceeded, models overfit the training data and the test loss starts to deteriorate across all of our test sets. We hypothesize that this size threshold corresponds to the capacity of the original back-translation model. This assertion suggests that in order for  back-translation to be beneficial for training large models, it has to be performed with a models with comparable capacity or higher. Although quite intriguing, we leave the verification of this hypothesis to future work. 

\begin{figure}[h]
  \centering
  \includegraphics[width=\textwidth]{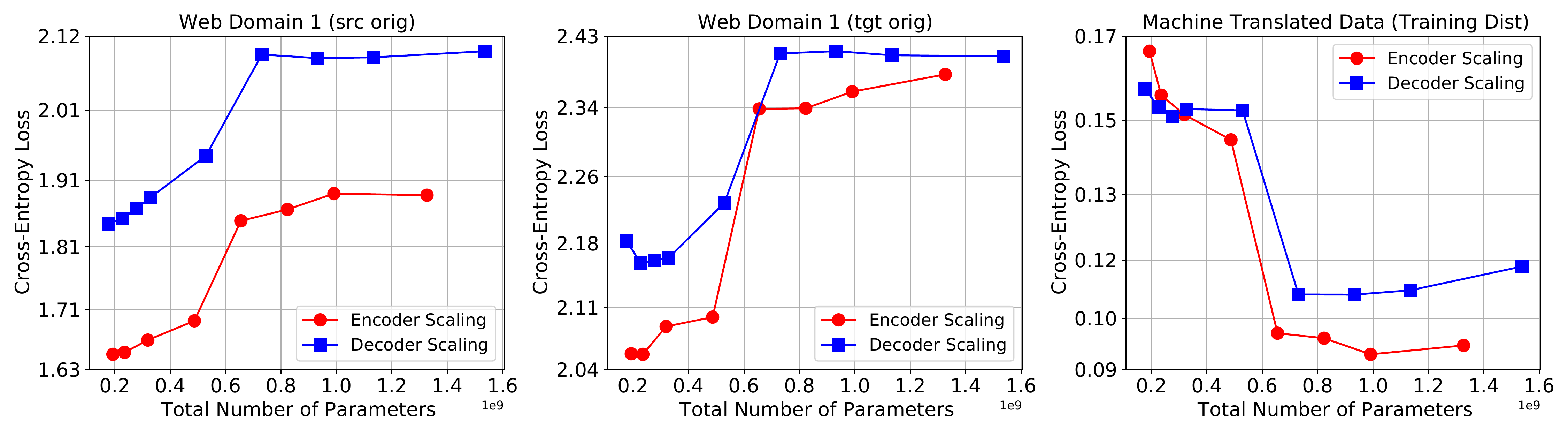}
  \vspace{-1.5em}
  \caption{Scaling behavior of models trained on forward translated data. Left / center: early stopping test loss on Web-Domain. Right: loss at the end of the training for a subset of the training data. \label{fig:FT_data}}
\end{figure}

Figure \ref{fig:FT_data} paints another interesting picture for the models trained on the source-original data only, implying the target side having the composition bias, expected to be simpler, dull and not rich in its content, in short - not natural looking. As experiments suggest, even our smallest models are able to achieve extremely low loss values (roughly $0.16$), with an apparent overfitting pattern. We believe the same phenomenon is also related to the "data simplification" effect seeked by non-autoregressive models in NMT \cite{zhou2021understanding}.

\section{Evolution of Generation Quality} \label{sec:gen_quality}

We examine the effects of scaling on the output quality as measured by BLEU score \footnote{We computed the BLEU scores using an internal reimplementation of Moses scorer: \url{mteval-v13a.pl}.}. For the analysis of this section, we focus on output generated via beam search \cite{wu2016google}. For tractability purposes, we do not attempt to tune the (many) hyper-parameters of beam-search for each model. Instead, we use the configuration optimized for the baseline model (listed in Appendix \ref{app:scaling_bleu}) in all the decoding tasks.

\begin{figure}[h]
  \centering
 \includegraphics[width=\textwidth]{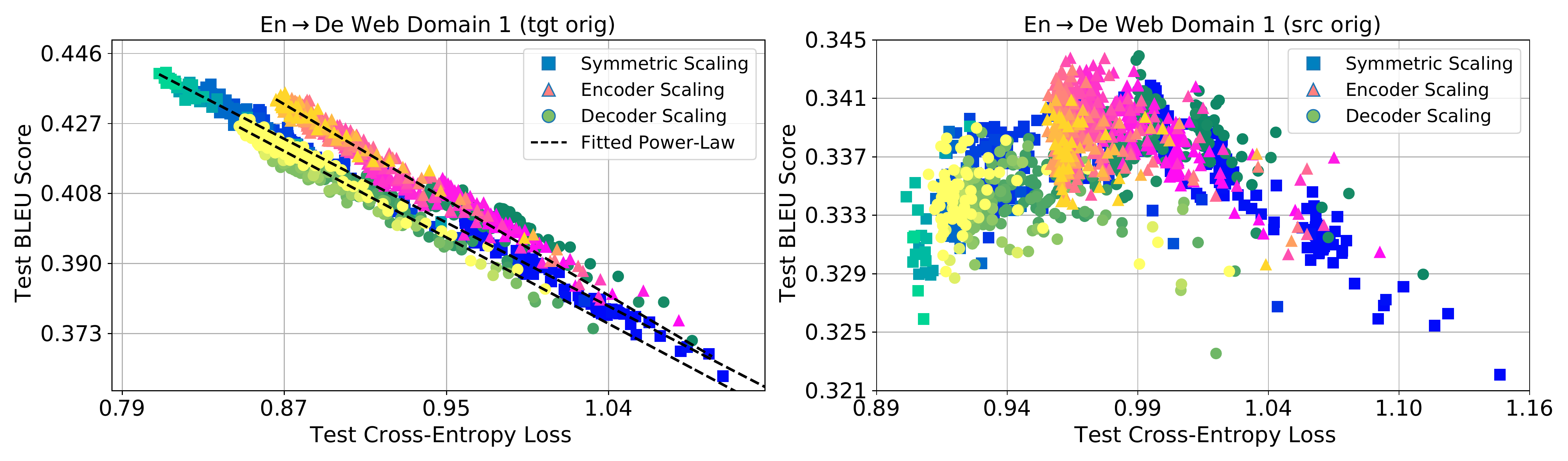}
 \includegraphics[width=\textwidth]{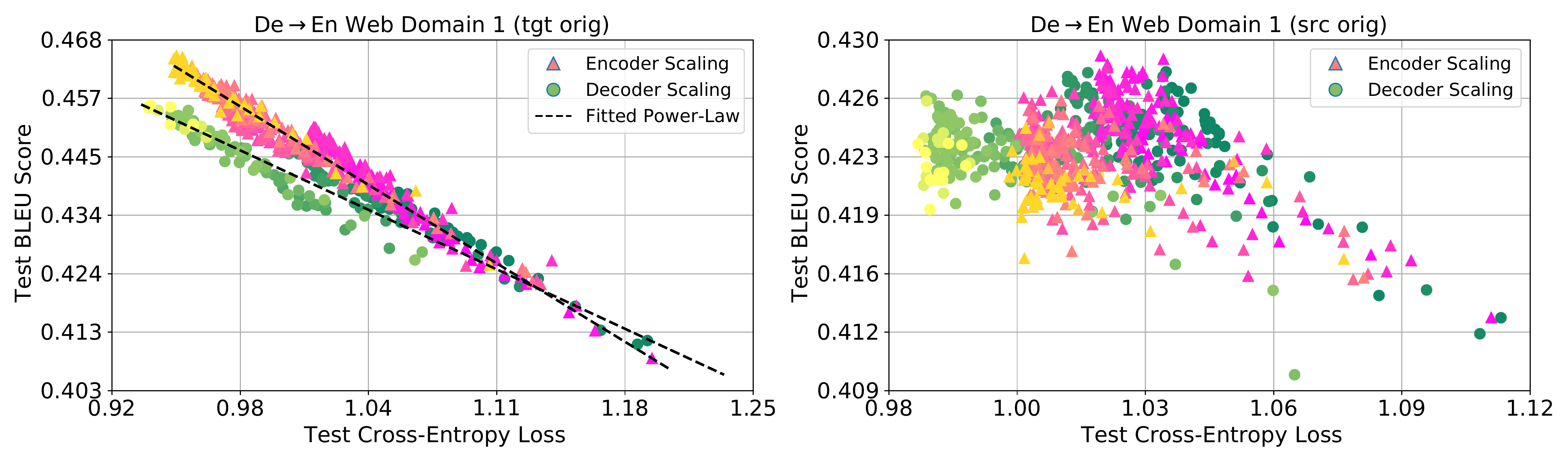}
  \caption{Log-log plot of the evolution of BLEU score as a function of cross-entropy loss for different models. For each scaling approach, warmer colors represent larger models. Each individual color represents different checkpoints of a single model during training. On target original data (left column), improvements to cross-entropy loss lead to consistent improvements in BLEU score. Dashed lines correspond to fit achieved by Eq. \eqref{eq:bleu_loss}. The relationship breaks down for source original data (right column). More examples of this phenomenon are presented in Appendix \ref{app:scaling_bleu}. \label{fig:bleu_loss}}
\end{figure}

Figure \ref{fig:bleu_loss} presents the co-evolution of BLEU score and cross-entropy loss throughout the training for all of our models. Depending on the construction of the test sets, two different empirical behaviors emerge. On target-original test sets, larger models are able to improve (lower) the test loss. These improvements in the loss are accompanied with consistent improvements (increases) in BLEU score. In fact, we observe that a simple power law of the form
\begin{equation} \label{eq:bleu_loss}
    \mbox{BLEU} = c_{B} L^{-p_{B}}, \qquad c_B,\; p_B > 0.
\end{equation}
can capture the relationship between BLEU score and cross-entropy loss for high-quality models. \footnote{We observe certain deviations from this trend for smaller models and for early checkpoints. We document these deviations in Appendix \ref{app:scaling_bleu}.}

In contrast, on source-original test sets, this relationship is absent; larger models consistently achieve better test losses, however, beyond a certain threshold, BLEU scores begin to deteriorate. Figures \ref{fig:bleu_loss_EnDe} and \ref{fig:bleu_loss_DeEn} exhibit that this phenomenon is not due to over-training; the BLEU score gap between large and small models is persistent throughout the training.

\begin{figure}[h]
  \centering
 \includegraphics[width=\textwidth]{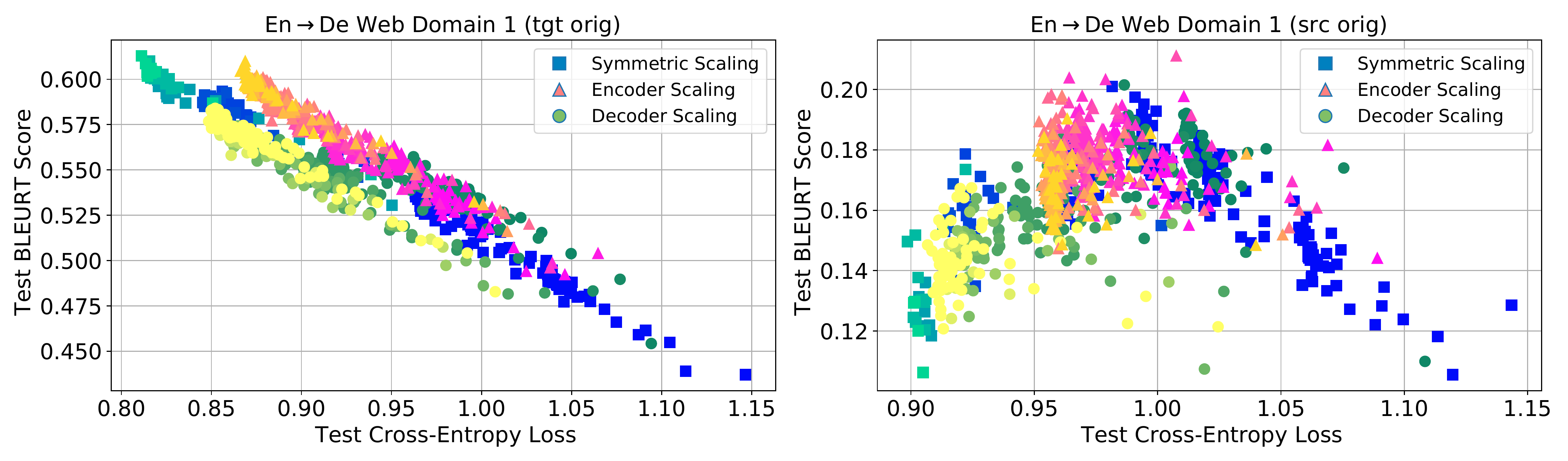}
 \includegraphics[width=\textwidth]{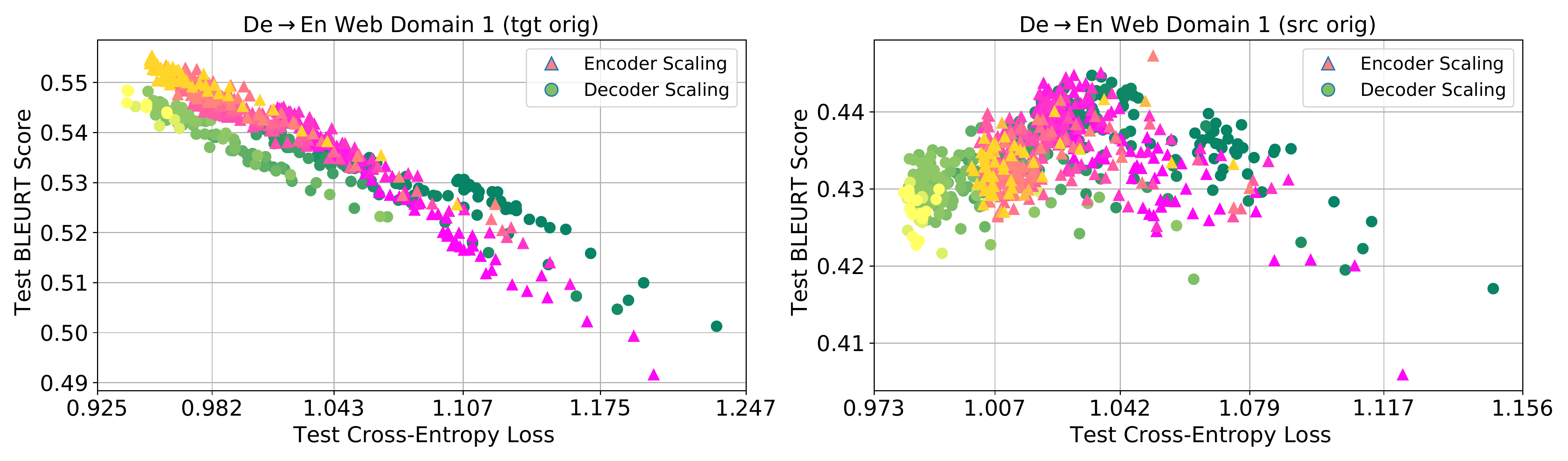}
  \caption{The evolution of BLEURT score as a function of cross-entropy loss for different models. For each scaling approach, warmer colors represent larger models. Each individual color represents different checkpoints of a single model during training. On target original data (left column), improvements to cross-entropy loss lead to consistent improvements in BLEURT score. The relationship breaks down for source original data (right column). More examples are provided in Appendix \ref{app:scaling_bleu}. \label{fig:bleurt_loss}}
\end{figure}

To ensure that this observation truly reflects the generation quality of the models (as opposed to potential biases of BLEU score), we repeat our analysis  with BLEURT score \cite{sellam2020bleurt, sellam2020learning}. The results are presented in Figure \ref{fig:bleurt_loss}. As the figure suggests, BLEURT scores closely mirror the behavior of BLEU scores with respect to model scaling. 

A careful look at the left-subplots of Figures \ref{fig:bleu_loss} \& \ref{fig:bleurt_loss} brings up another interesting trend. At similar values of the test loss, encoder-scaled models result in better generation quality compared to decoder-scaled models. This findings agrees with previous work that relied on encoder-scaling when optimizing for BLEU and inference latency \citep{kasai2021deep}. Whether these differences in the effects of encoder-scaling and decoder-scaling are caused by insufficient search algorithms, or just  different model fits from different architectural priors is left to future work.

\section{Conclusion and Limitation}

In this work we have attempted to quantify the evolution of model quality as a function of model capacity for encoder-decoder NMT models.

While a univariate scaling law describing the cross-entropy as a function of the total number of parameters in the model is insufficient, a bivariate law treating the number of  encoder and decoder parameters as separate variables adequately describes the scaling behavior of these models under various scaling strategies. We validate this behavior on 2 language pairs and on a variety of evaluation sets with different compositions. Whether this behavior is intrinsic to the encoder-decoder architecture, or arising from the nature of the NMT task, requires further study.

Next, we demonstrate that this scaling behavior is highly dependent on the composition of the evaluation data, specifically on whether the source or the target sentences are ``original''. Our findings indicate that target-original evaluation sets continue benefiting from model scaling throughout our range of measurements, while the reducible error on source-original evaluation sets quickly saturates to $0$. This could be an artifact of the lack of diversity in translated text; a simpler target distribution doesn't require much capacity to model while generating fluent or natural-looking text could benefit much more from scale.

We also study how the composition of training data affects the scaling behavior of models. When training on target-original (back-translated) text, model quality keeps improving until a point after which the trend saturates. In our study the capacity where saturation manifests first is perilously close to the capacity of the model used for back-translation, indicating that the capacity of the generative model used to generate synthetic text might have a role to play, but this requires further investigation. When training on source-original text, even low-capacity models are sufficient to reach the irreducible loss region, painting a gloomy picture for synthetic data. While we have explored these ideas in the context of machine translation, given the proliferation of generative models this problem will likely be a challenge for future practitioners training on web-scraped monolingual datasets as well. For low-resource languages, the proliferation of machine translated text is already a problem given that a significant portion of web text in these languages is machine translated.

Finally, we attempt to understand how generation quality evolves with the improvements in cross-entropy resulting from model scaling. As with our previous findings, dataset composition plays a major role in determining the trends. For source-original evaluation sets, the correlation between cross-entropy and generation quality breaks down. On target-original evaluation, we observe an inverse correlation between cross-entropy and BLEU/BLEURT, suggesting that improved model fit results in a corresponding improvement in generation quality. The slope of this relationship is different for encoder-scaling and decoder-scaling, with encoder-scaled models performing better on BLEU/BLEURT than decoder-scaled models, at the same level of cross-entropy loss. Whether this is an artifact of our search strategy (beam search, tuned to a 6L-encoder 6L-decoder model) or the difference in architectural priors is something that requires further investigation.

Our findings suggest that scaling behavior of encoder-decoder NMT models is predictable, but the exact formulation of scaling laws might vary depending on the particular architecture or task being studied. Our empirical findings also raise concerns regarding the effect of synthetic data on model scaling and evaluation, and how proliferation of machine generated text might hamper the quality of future models trained on web-text.

\begin{ack}
We would like to thank the Google Translate and Google Brain teams for their useful input and discussion, and the entire Lingvo development team for their foundational contributions to this project. We would also like to thank Kyunghyun Cho, Ethan Dyer, Yasaman Bahri and David Luan for their insightful comments.
\end{ack}

\bibliographystyle{plain}
\bibliography{main}

\newpage
\appendix

\section{Architecture and Hyper-Parameter Details}
\label{app:hyps}
As described in Section \ref{sec:exps}, all our models use a similar configuration for their Transformer Blocks. In particular, we fix model dimension to $1024$, feed-forward layer dimension to $8192$, number of attention heads to $16$, and attention hidden dimension to $1024$. Our models use a sentence-piece vocabulary of size $32000$.  

\paragraph{Regularization:} We use a dropout of $0.1$ for residuals, feed-forward activations and attention. Models are trained with label smoothing of magnitude $0.1$. To improve the training stability, all models use logit clipping of $10$.

\paragraph{Optimizer:} We use Adafactor \cite{shazeer2018adafactor} optimizer for training our models. We use $40$k linear warm-up steps and an inverse square root learning rate schedule. For Adafactor we used momentum with $0.9$ and factored second moment to save memory.

Table \ref{tab:encoder_models} (and Table \ref{tab:decoder_models} resp.) describes the parameter decomposition of the encoder scaling (decoder scaling) models. Table \ref{tab:sym_models} describes the parameter counts for the symmetric scaling models. The largest model we used (64L-64L) has more than $3$ billion parameters while the smallest model we used (2L-2L) has only $92$M non-embedding parameters. 

\begin{table}[h!]
  \caption{Parameter decomposition of the encoder scaling models. The total number of parameters includes $98$M parameters representing the softmax and embedding layers.}
  \label{tab:encoder_models}
  \centering
  \begin{tabular}{crccr}
    \toprule
    \multicolumn{2}{c}{Encoder} 
    &
    \multicolumn{2}{c}{Decoder} \\ \cmidrule(r){1-2}\cmidrule(l){3-4}
    Layers & Parameters & Layers & Parameters & Total Parameters\\
    \midrule
    $2$    & $42$M   & $6$ &  $151$M &  $291$M  \\
    $4$    & $84$M   & $6$ &  $151$M &  $333$M  \\
    $8$    & $168$M  & $6$ &  $151$M &  $417$M  \\
    $12$   & $252$M  & $6$ &  $151$M &  $501$M  \\
    $16$   & $336$M  & $6$ &  $151$M &  $585$M  \\
    $20$   & $420$M  & $6$ &  $151$M &  $669$M  \\
    $24$   & $504$M  & $6$ &  $151$M &  $753$M  \\
    $28$   & $588$M  & $6$ &  $151$M &  $837$M  \\
    $32$   & $672$M  & $6$ &  $151$M &  $921$M  \\
    $36$   & $756$M  & $6$ &  $151$M & $1005$M  \\
    $40$   & $840$M  & $6$ &  $151$M & $1089$M  \\
    $48$   & $1007$M & $6$ &  $151$M & $1257$M  \\
    $56$   & $1175$M & $6$ &  $151$M & $1425$M  \\
    $64$   & $1343$M & $6$ &  $151$M & $1593$M  \\
    \bottomrule
  \end{tabular}
\end{table}

\begin{table}[h!]
  \caption{Parameter decomposition of the decoder scaling models. The total number of parameters includes $98$M parameters representing the softmax and embedding layers. Note that the 6L-6L model is the baseline model we used for hyper-parameter tuning.}
  \label{tab:decoder_models}
  \centering
  \begin{tabular}{cccrr}
    \toprule
    \multicolumn{2}{c}{Encoder} 
    &
    \multicolumn{2}{c}{Decoder} \\ \cmidrule(r){1-2}\cmidrule(l){3-4}
    Layers & Parameters & Layers & Parameters & Total Parameters\\
    \midrule
    $6$ & $126$M  &  $2$  & $50$M   & $275$M  \\
    $6$ & $126$M  &  $4$  & $101$M  & $325$M  \\
    $6$ & $126$M  &  $6$  & $151$M  & $375$M  \\
    $6$ & $126$M  &  $8$  & $202$M  & $426$M  \\
    $6$ & $126$M  &  $12$ & $302$M  & $527$M  \\
    $6$ & $126$M  &  $16$ & $403$M  & $627$M  \\
    $6$ & $126$M  &  $20$ & $504$M  & $728$M  \\
    $6$ & $126$M  &  $24$ & $605$M  & $829$M  \\
    $6$ & $126$M  &  $28$ & $705$M  & $930$M  \\
    $6$ & $126$M  &  $32$ & $806$M  & $1030$M  \\
    $6$ & $126$M  &  $36$ & $907$M  & $1131$M  \\
    $6$ & $126$M  &  $40$ & $1008$M & $1232$M  \\
    $6$ & $126$M  &  $48$ & $1209$M & $1433$M  \\
    $6$ & $126$M  &  $56$ & $1411$M & $1635$M  \\
    $6$ & $126$M  &  $64$ & $1612$M & $1836$M  \\
    \bottomrule
  \end{tabular}
\end{table}

\begin{table}[h!]
  \caption{Parameter decomposition of the symmetric scaling models trained for English$\to$German translation task. The total number of parameters includes $98$M parameters representing the softmax and embedding layers.}
  \label{tab:sym_models}
  \centering
  \begin{tabular}{crcrr}
    \toprule
    \multicolumn{2}{c}{Encoder} 
    &
    \multicolumn{2}{c}{Decoder} \\ \cmidrule(r){1-2}\cmidrule(l){3-4}
    Layers & Parameters & Layers & Parameters & Total Parameters\\
    \midrule
    $2$  & $42$M  &  $2$  & $50$M   & $191$M  \\
    $3$  & $63$M  &  $3$  & $76$M  & $237$M  \\
    $4$  & $84$M  &  $4$  & $101$M  & $283$M  \\
    $5$  & $105$M  &  $5$  & $126$M  & $329$M  \\
    $8$  & $168$M  &  $8$  & $202$M  & $468$M  \\
    $16$ & $336$M  &  $16$ & $403$M  & $837$M  \\
    $20$ & $420$M  &  $20$ & $504$M  & $1022$M  \\
    $24$ & $504$M  &  $24$ & $605$M  & $1207$M  \\
    $32$ & $672$M  &  $32$ & $806$M  & $1576$M  \\
    $48$ & $1007$M  &  $48$ & $1209$M  & $2315$M  \\
    $56$ & $1175$M  &  $56$ & $1411$M  & $2684$M  \\
    $64$ & $1343$M  &  $64$ & $1612$M & $3054$M  \\
    \bottomrule
  \end{tabular}
\end{table}

\clearpage
\section{Scaling Laws for Other Test Sets}
\label{app:full_results}

In order to keep the discussion in the main text focused, we only presented scaling laws for Web Domain 1 test sets. These test sets were chosen as they had a domain similar to the training data (i.e. web). In this appendix, we repeat the same analysis for our other test sets. The details of these test sets are described in Section \ref{sec:exps}.

\begin{figure}[h!]
  \centering
  \includegraphics[width=0.95\textwidth]{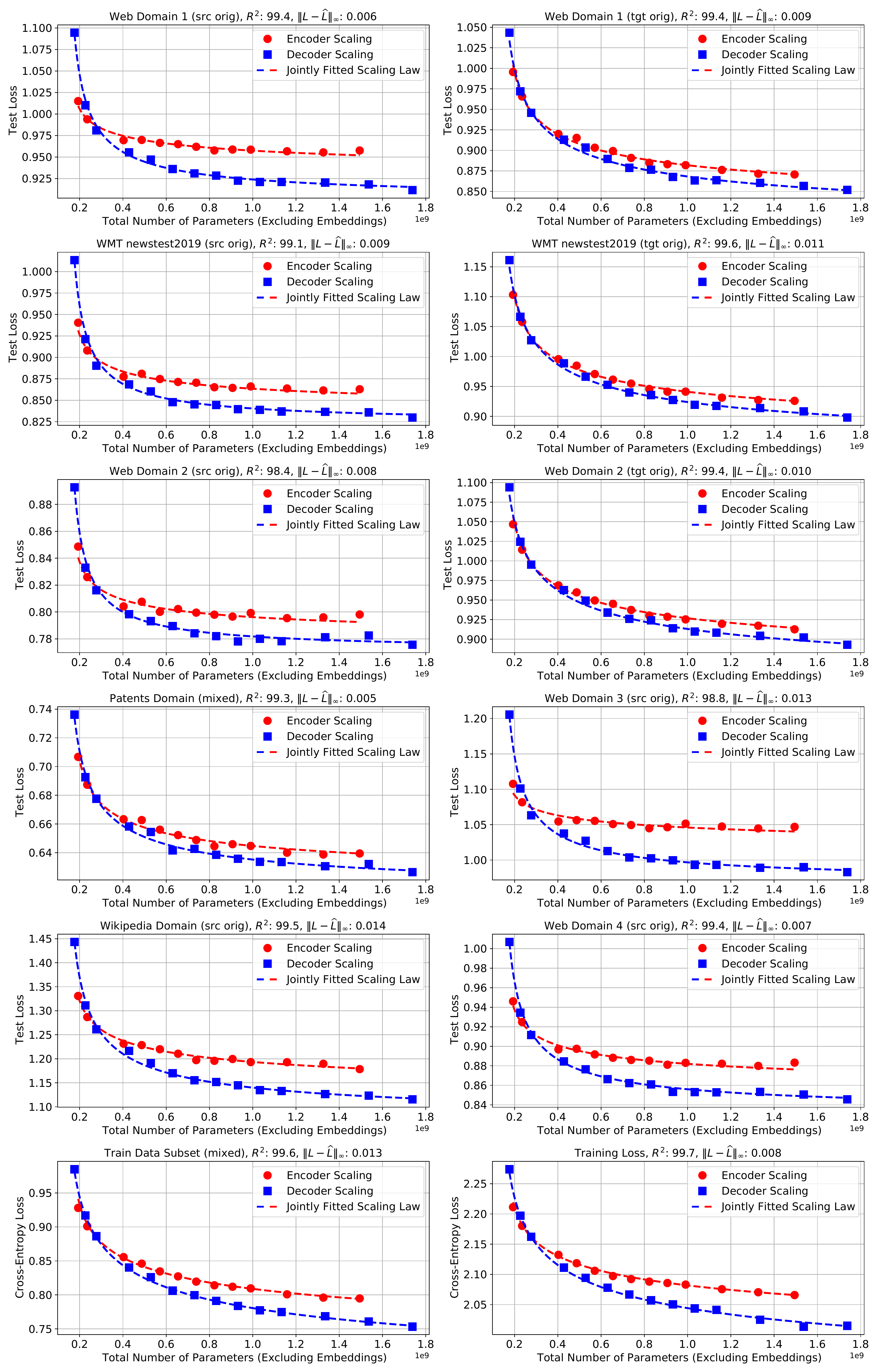}
  \caption{Fitted scaling law (Eq. \eqref{eq:scaling_law}) for English$\to$German translation task. The scaling law captures the scaling behavior of the models over a diverse collection of test sets and domains. The last row describes the evolution of the cross-entropy loss on the training data (with and without regularization effect). \label{fig:EnDe_all}}
\end{figure}

Figure \ref{fig:EnDe_all} demonstrates how well the scaling law in Eq. \eqref{eq:scaling_law} fits the empirical scaling behavior of our models on all our test sets. For each test set, we have fitted the law jointly on the final test loss achieved by the encoder and decoder scaling models. We measure the final test loss by the median test loss over steps $450$K to $500$K. Details of the fitting procedure are provided in Appendix \ref{app:curve_fitting}.

The results suggest that Eq. \eqref{eq:scaling_law} is closely capturing the scaling behavior of the model for all the test sets / domains. In the last row, we also demonstrate the fit for the training data (left column) and the training loss (cross entropy on training data plus regularization, right column). We observe that the scaling law is almost perfectly fitting the empirical data in these cases.

\begin{figure}[h!]
  \centering
  \includegraphics[width=0.95\textwidth]{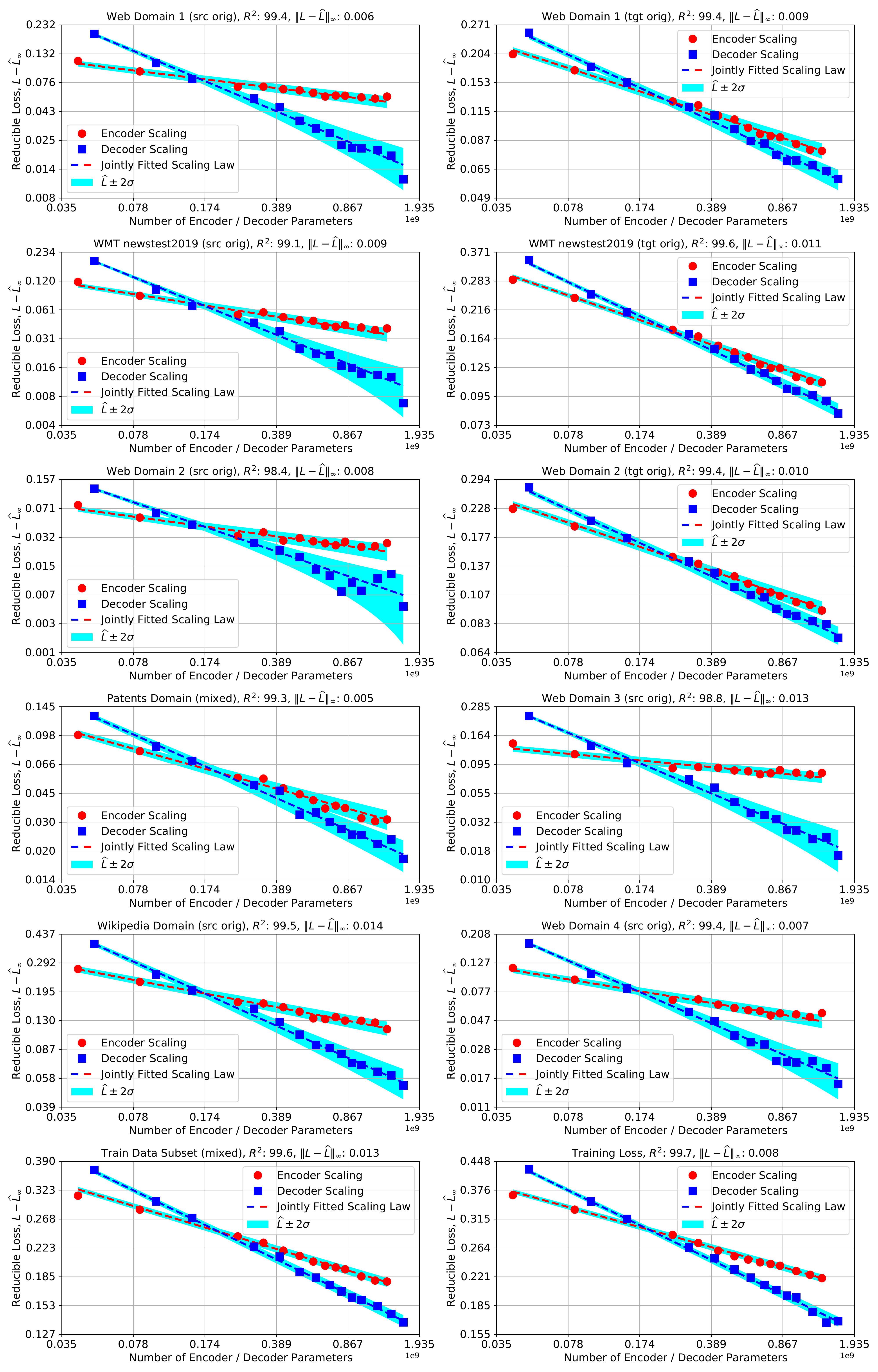}
  \caption{Fitted scaling law (Eq. \eqref{eq:scaling_law}) for English$\to$German translation task. Here, we use a log-log plot in order to inspect the fit more closely. Shaded cyan regions correspond to the uncertainty region given by $\pm 2 \times$standard deviation. Per test set standard deviations are provided in Table \ref{tab:variations}. \label{fig:EnDe_all_loglog}}
\end{figure}

To examine the fit more closely, in Figure \ref{fig:EnDe_all_loglog}, we have plotted the same data but with several modifications: 

\begin{enumerate}
    \item Instead of plotting the final loss, we plot the \emph{reducible} component of the final loss ($L - L_{\infty}$). As the true value of $L_{\infty}$ is unknown, we use the value given by the fit of the scaling law. 
    
    \item For encoder scaling models, we plot the (reducible) loss against the number of encoder parameters (as opposed to the total number of parameters). Similarly, for decoder scaling models, we plot the loss against the number of decoder parameters.

    \item We use log-log scaling on the axes.
    
    \item We use the results of Table \ref{tab:variations} to provide a confidence region around our predictions. This confidence region quantifies our expected uncertainty caused by randomness in the initialization and training pipeline. 
\end{enumerate}

Eq. \eqref{eq:scaling_law} predicts that the relationship between the empirical final loss values from the encoder (decoder) scaling models and the number of encoder (decoder) parameters should appear linear on these plots. Figure \ref{fig:EnDe_all_loglog} suggests that the empirical scaling behavior of these models conforms closely to these predictions.

\begin{figure}[h!]
  \centering
  \includegraphics[width=\textwidth]{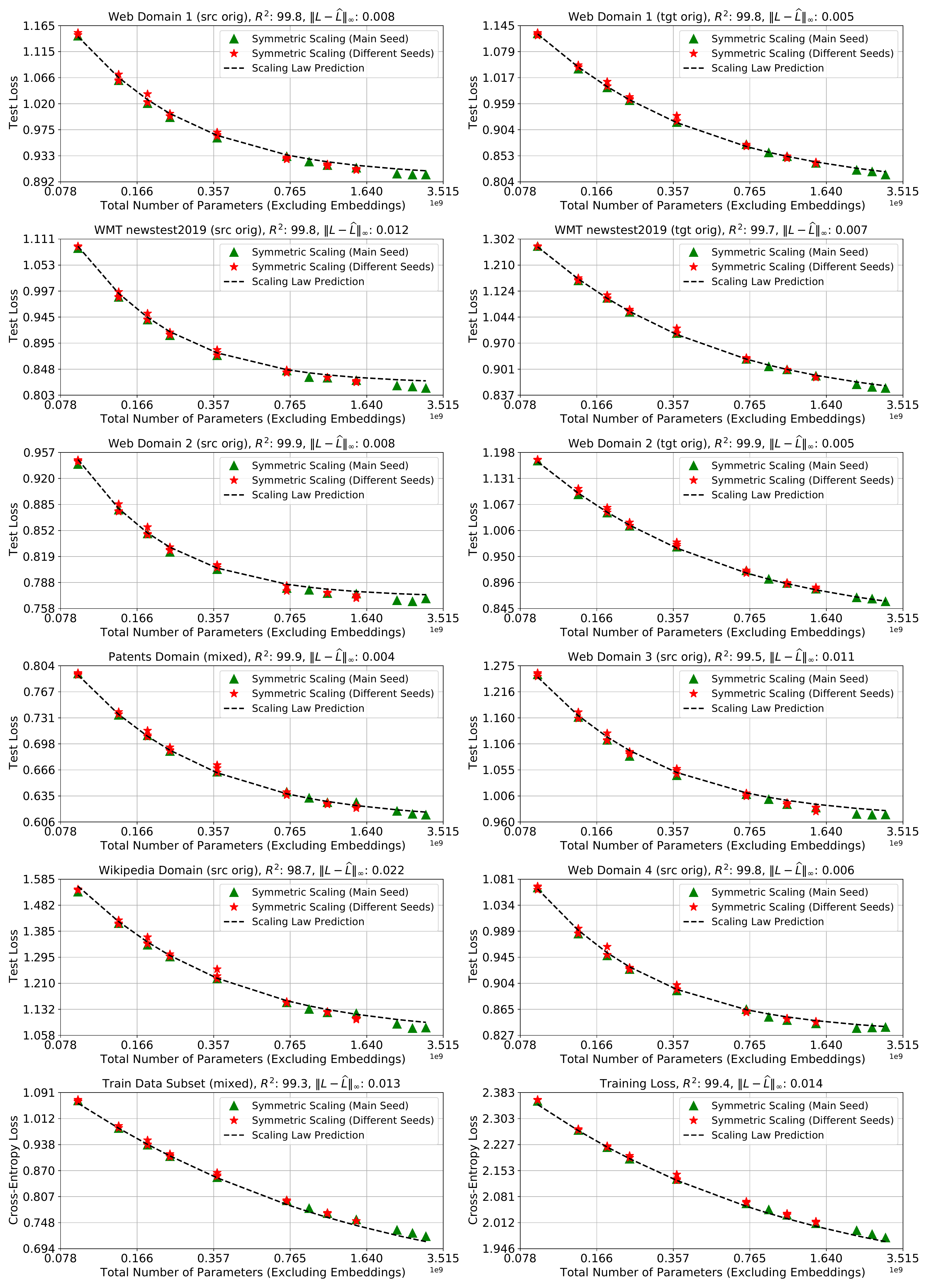}
  \caption{Out-of-sample prediction accuracy of English$\to$German scaling laws on symmetric scaling models. Scaling laws are fitted only using the encoder and decoder scaling models. Nevertheless, they accurately predict the scaling behavior of symmetric scaling models. \label{fig:EnDe_all_OOS}}
\end{figure}

To see if Eq. \eqref{eq:scaling_law} continues to capture the scaling behavior of the models out-of-sample, we compare the predictions of the scaling law with the empirical loss values for symmetric scaling models. In other words, we examine how well the scaling laws fitted only using encoder / decoder scaling models predict the final test  loss achieved by symmetric scaling models of different sizes. Figure \ref{fig:EnDe_all_OOS} shows this comparison for all of our test sets. We observe a remarkable match between the predictions of the scaling law and the empirical loss values across the board. These observations confirm that Eq. \eqref{eq:scaling_law} is able to capture the scaling behavior of the model regardless of the scaling approach.

\begin{figure}[h!]
  \centering
  \includegraphics[width=0.97\textwidth]{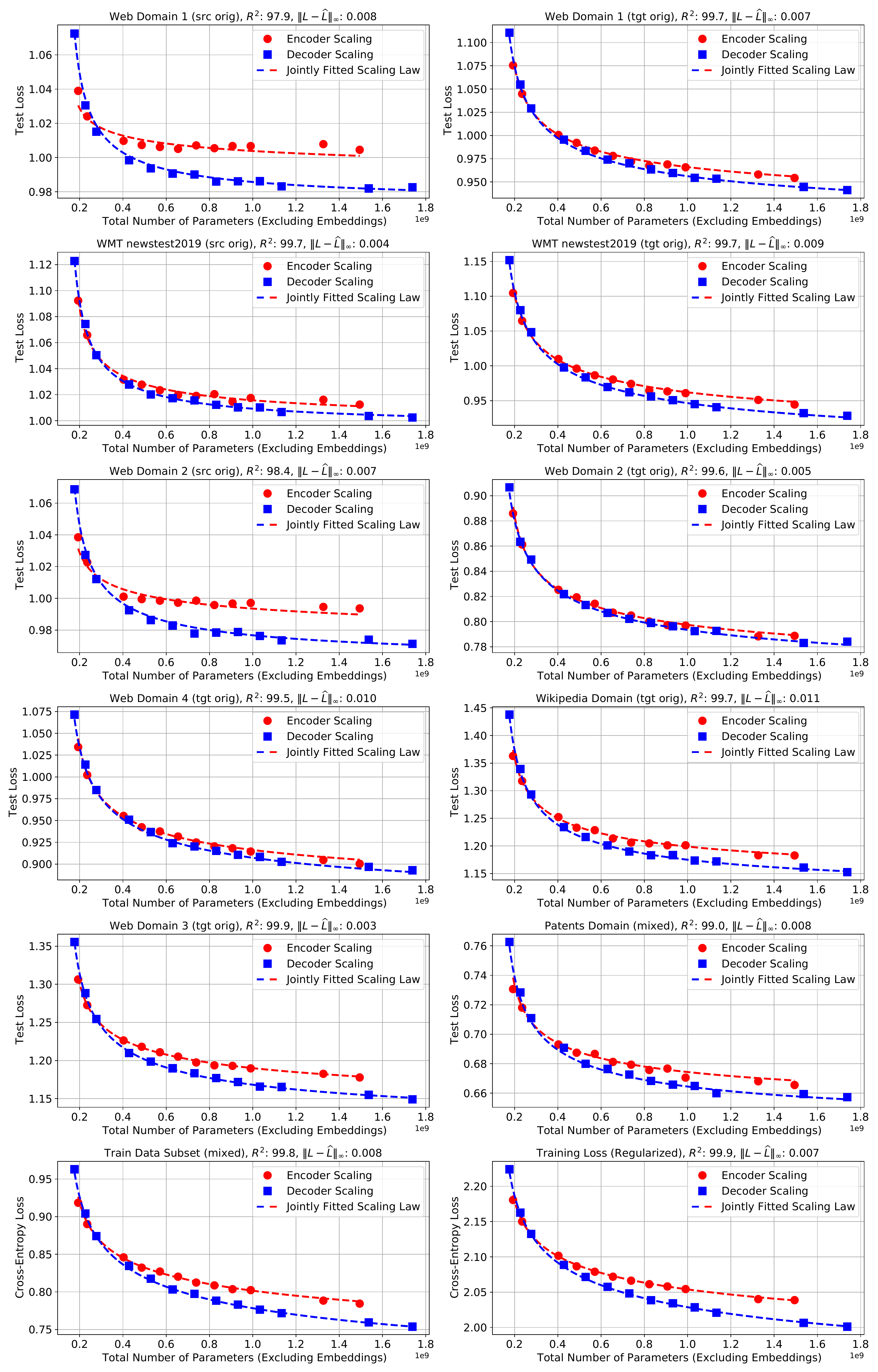}
  \caption{Fitted scaling law (Eq. \eqref{eq:scaling_law}) for German$\to$English translation task.  The scaling law captures the scaling behavior of the models over a diverse collection of test sets and domains. \label{fig:DeEn_all}}
\end{figure}

\begin{figure}[h!]
  \centering
  \includegraphics[width=\textwidth]{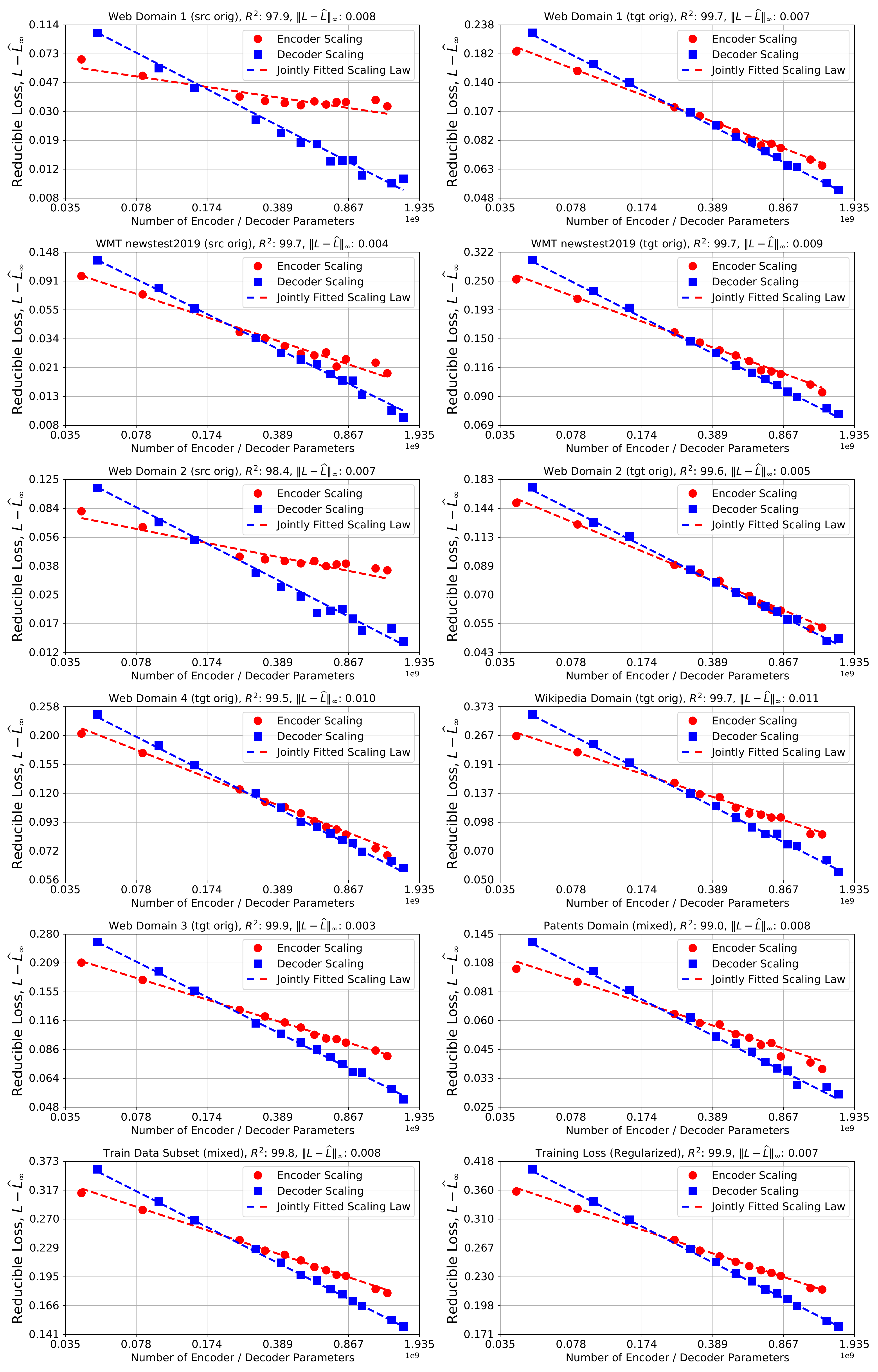}
  \caption{German$\to$English scaling law fits on log-log scale. \label{fig:DeEn_all_loglog}}
\end{figure}

\clearpage
\section{Quantifying the Random Variations in the Results} 
\label{app:variation}

\begin{figure}[h!]
  \centering
  \includegraphics[width=\textwidth]{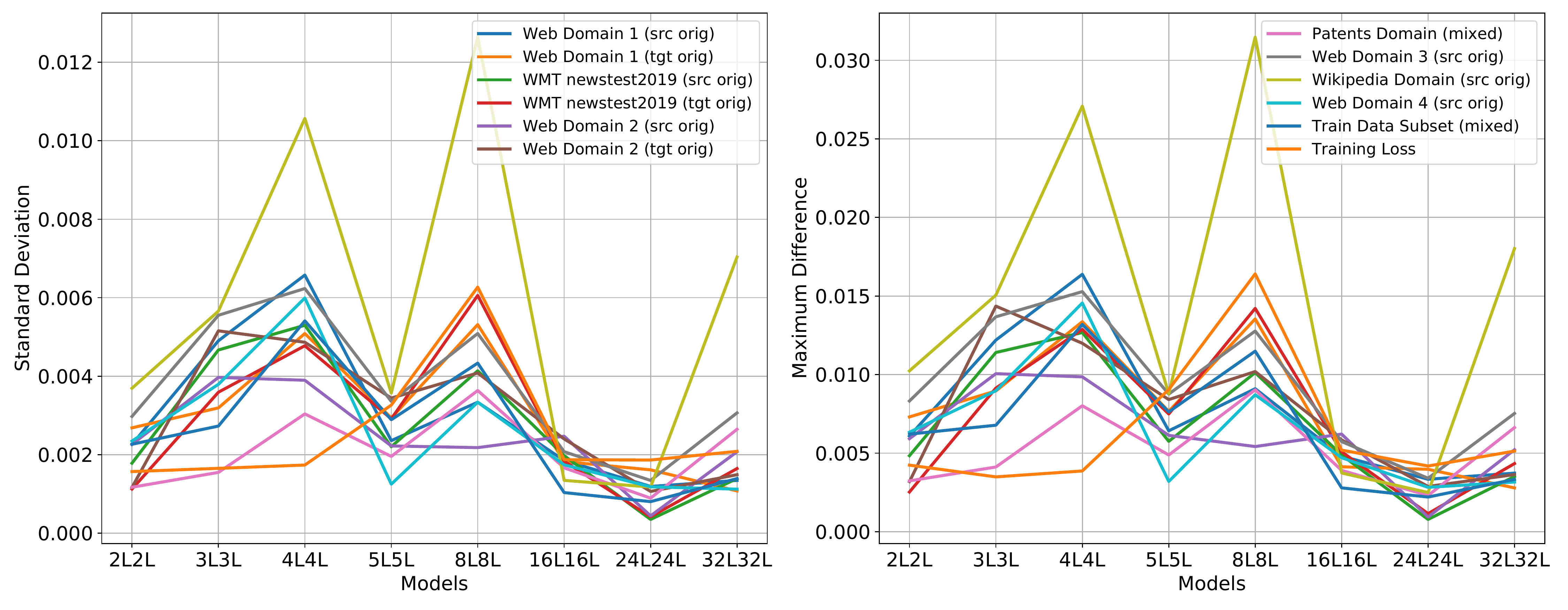}
  \caption{Variability of the final test loss across four different seeds. \label{fig:variation}}
\end{figure}

Note that the final test loss achieved by the model is a random quantity. Randomness is incorporated into the training pipeline through the initialization step, data order, and hardware failures / preemptions. To quantify the magnitude of the fluctuations caused by this randomness, we retrain a subset of our models (2L-2L, 3L-3L, 4L-4L, 5L-5L, 8L-8L, 16L-16L, 24L-24L, and 32L-32L) with $4$ different seeds. Figure \ref{fig:variation} presents standard deviation (left) and maximum difference (right) of the final test loss values observed for each model.  

\begin{table}[h!]
  \caption{Variability of final test loss for each test dataset (averaged over all models).}
  \label{tab:variations}
  \centering
  \begin{tabular}{lcc}
    \toprule
    Dataset & Average Standard Deviation & Average Maximum Deviation \\
    \midrule
    Web Domain 1 (src orig)     &  $0.0030$  & $0.0078$\\
    Web Domain 1 (tgt orig)     &  $0.0030$  & $0.0077$\\
    \midrule
    WMT newstest2019 (src orig) & $0.0027$   & $0.0067$\\
    WMT newstest2019 (tgt orig) & $0.0028$   & $0.0071$\\
    \midrule
    Web Domain 2 (src orig)     & $0.0024$   & $0.0062$\\
    Web Domain 2 (tgt orig)     & $0.0030$   & $0.0076$\\
    \midrule
    Patents Domain (mixed)      & $0.0021$   & $0.0053$\\
    Web Domain 3 (src orig)     & $0.0037$   & $0.0094$\\
    Wikipedia Domain (src orig) & $0.0057$   & $0.0146$\\
    Web Domain 4 (src orig)     & $0.0026$   & $0.0066$\\
    \midrule
    Train Data Subset (mixed)   & $0.0026$   & $0.0067$\\
    Training Loss               & $0.0025$   & $0.0064$\\
    \bottomrule
  \end{tabular}
\end{table}

\clearpage
\section{Proofs} 
\label{app:proofs}
\subsection{Proof of proposition \ref{prop:optimal_scaling}}
\begin{proof}
Let $\beta \equiv \alpha \bar{N}_e^{p_e} \bar{N}_d^{p_d}$. Then the optimal encoder / decoder sizes are optimal parameters of the following optimization problem:
\begin{align}
    \begin{array}{cc}
      \mbox{minimize }_{N_e, N_d}   &  \beta N_e^{-p_e} N_d^{-p_d} \\
       \mbox{s.t.}  & N_e + N_d \leq B
    \end{array}.
\end{align}
To convert the problem to a convex problem, we instead consider the log of the objective and adopt the following change of variables:
\begin{align}
    u \equiv \log(N_e), \qquad v \equiv \log(N_d).
\end{align}
The transformed optimization problem is of the form:
\begin{align} \label{prob:cvx}
    \begin{array}{cc}
      \mbox{minimize }_{u, v}   &  -p_e u - p_d v \\
       \mbox{s.t.}  & \exp(v) + \exp(u) \leq B
    \end{array}.
\end{align}
Note that \eqref{prob:cvx} is now convex and therefore, we can use KKT conditions to solve for the optimum. The Lagrangian has the form:
\begin{align}
    \mathcal{L}(u, v, \lambda) = -p_e u - p_d v + \lambda \bigg(B - \exp(v) - \exp(u)\bigg).
\end{align}
Solving for the first-order conditions yield:
\begin{align}
    -p_e &= \lambda \exp(u^*) \\
    -p_d &= \lambda \exp(v^*).
\end{align}
Since the constraint is binding, $\lambda \not = 0$. Therefore, we can divide both sides of the equations above which yields:
\begin{align} \label{eq:optimal_ratio}
    \frac{p_e}{p_d} &= \frac{\exp(u^*)}{\exp(v^*)} = \frac{N_e^*}{N_d^*}.
\end{align}
Substituting \eqref{eq:optimal_ratio} in the constraint yields:
\begin{align} \label{eq:optimal_params_v2}
    N_e^* = \frac{p_e}{p_e + p_d} B, \qquad N_d^* = \frac{p_d}{p_e + p_d} B.
\end{align}
Finally, we substitute \eqref{eq:optimal_params_v2} in the scaling law which yields:
\begin{align}
    \hat{L}_{opt}(B) &= \alpha \bigg( \frac{\bar{N}_e (p_e + p_d)}{p_e B} \bigg)^{p_e} \bigg(\frac{\bar{N}_d (p_e + p_d)}{p_d B} \bigg)^{p_d} + L_{\infty} \\
                     &= \alpha \bigg( \frac{\bar{N}_e (p_e + p_d)}{p_e} \bigg)^{p_e} \bigg(\frac{\bar{N}_d (p_e + p_d)}{p_d} \bigg)^{p_d} B^{-(p_e + p_d)}+ L_{\infty}
\end{align}
\end{proof}

\newpage
\section{Curve Fitting Details}
\label{app:curve_fitting}
We use \lstinline[language=Python,style=mystyle]{scipy.optimize.least_squares} function for curve fitting throughout this paper \footnote{\url{https://docs.scipy.org/doc/scipy/reference/generated/scipy.optimize.least_squares.html}}. To have some robustness to outliers, we use the \lstinline[language=python]{loss=`soft_l1'} option which is a popular option for robust regression. The code snippet below shows the exact arguments we use for fitting the scaling laws:

\begin{lstlisting}[language=Python]
def func(p, x, y):
  """Fitting a bivariate scaling law.
  
  p: A 1-D array of dim 4, corresponding to alpha, p_e, p_d, c.
  x: A matrix of dimension n \times 2. First column encoder params, second col decoder params.
  y: A 1-D array of log-pplx of dim n."""
  x_e = NE_bar / x[:, 0]
  x_d = ND_bar / x[:, 1]
  return p[0] * np.power(x_e , p[1]) * np.power(x_d , p[2]) + p[3] - y    

def fit_model(x, y, f_scale):
  X = x.to_numpy().copy()
  y = y.to_numpy().copy()
  if np.isnan(X).any() or np.isnan(y).any():
    raise ValueError('Data contains NaNs')
  if len(y.shape) > 1 or y.shape[0] != X.shape[0]:
    raise ValueError('Error in shapes')

  p0 = np.zeros((4,))
  p0[0] = 0.2 # alpha
  p0[1] = 0.4 # p_e
  p0[2] = 0.6 # p_d
  p0[3] = 1.0 # c
  fit = least_squares(func, p0, loss='soft_l1', f_scale=f_scale,
                       args=(X, y), max_nfev=10000, bounds=(0, 10))
  return fit
\end{lstlisting}

The \lstinline[language=python]{`soft_l1'} loss chosen above applies $\ell_2$ penalty on small residuals and a $\ell_1$-like penalty on outlier residuals. The argument \lstinline[language=python]{f_scale} determines the boundary where the transition between the two different behaviors occur. For the results presented in this paper, we choose \lstinline[language=python]{f_scale} from the grid given by \lstinline[language=python]{np.geomspace(0.001, 0.025, num=25)}. Choosing \lstinline[language=python]{f_scale=0.025} effectively yields a least-squares regression while smaller values add more robustness to outliers. 
\newpage
\section{Analysis of the Generation Quality}
\label{app:scaling_bleu}

\paragraph{Decoding:} As described in Section \ref{sec:gen_quality}, we use beam-search for decoding \cite{wu2016google}. To keep the experiments tractable, we did not attempt to tune the hyper-parameters of beam-search for each model. Instead, we use the same hyper-parameters (optimized for the baseline model) for all our decoding jobs. In particular, we fix the length normalization parameter to $1.0$ and number of beams to $4$. 

\begin{figure}[h]
  \centering
  \includegraphics[width=\textwidth]{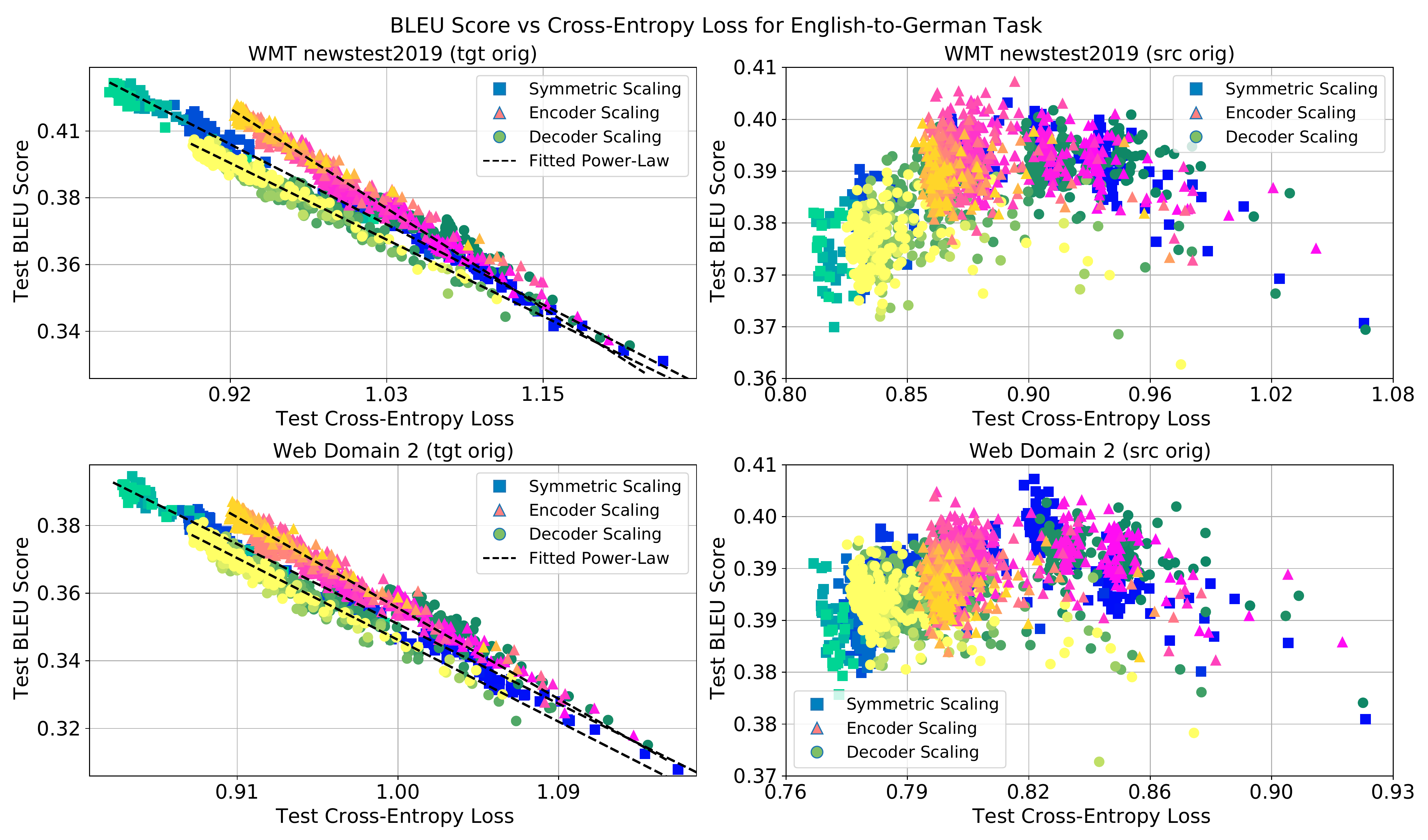}
  \includegraphics[width=\textwidth]{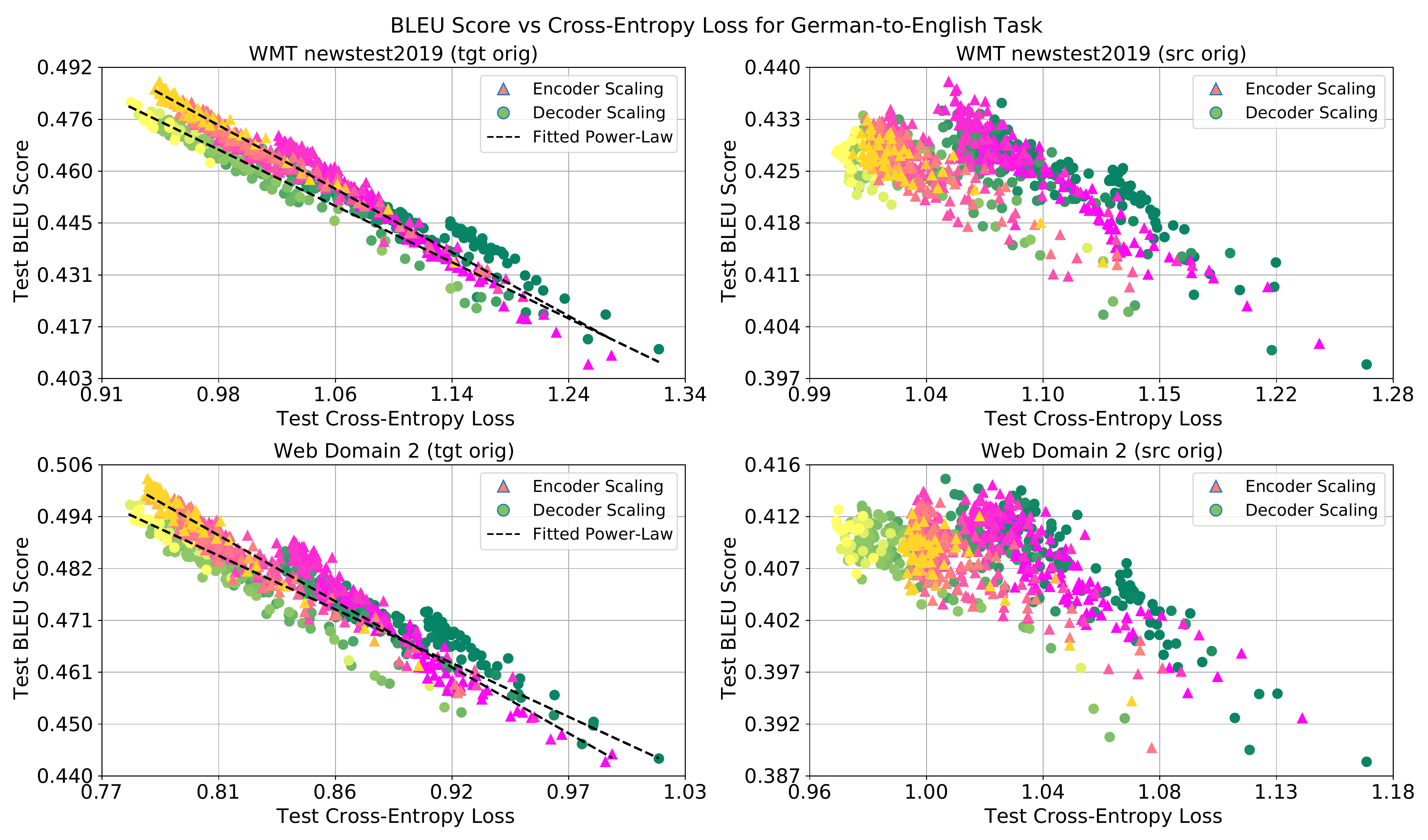}
  \caption{Log-log plot of the evolution of BLEU score as a function of cross-entropy loss for different models. For each scaling approach, warmer colors represent larger models. Each individual color represents different checkpoints of a single model during training. On target original data (left column), improvements to cross-entropy loss lead to consistent improvements in BLEU score. Dashed lines correspond to fit achieved by Eq. \eqref{eq:bleu_loss}. The relationship breaks down for source original data (right column). \label{fig:bleu_loss_app}}
\end{figure}

\paragraph{BLEU-Cross Entropy Loss Co-Evolution:} Figure \ref{fig:bleu_loss_app} presents the relationship between BLEU score and cross-entropy loss for various test datasets. The results closely mimic the phenomenon observed in Figure \ref{fig:bleu_loss}: On target original data, improvements to cross-entropy loss are accompanied with improvements in BLEU score. On source original data however, beyond a certain point, cross-entropy loss and BLEU score exhibit diverging behaviors.

We observe that in large well-trained models, the relationship between BLEU and cross-entropy loss on target-original data is well captured by the power law presented in Eq. \eqref{eq:bleu_loss}. The fit achieved by this power law is plotted in our figures. We observe that fitted power laws for encoder scaling models consistently attain larger exponents compared to decoder or symmetrically scaled models. This reflects the fact that encoder scaling models are more successful in improving the generation quality (as measured by BLEU).  

\begin{figure}[h]
  \centering
  \includegraphics[width=\textwidth]{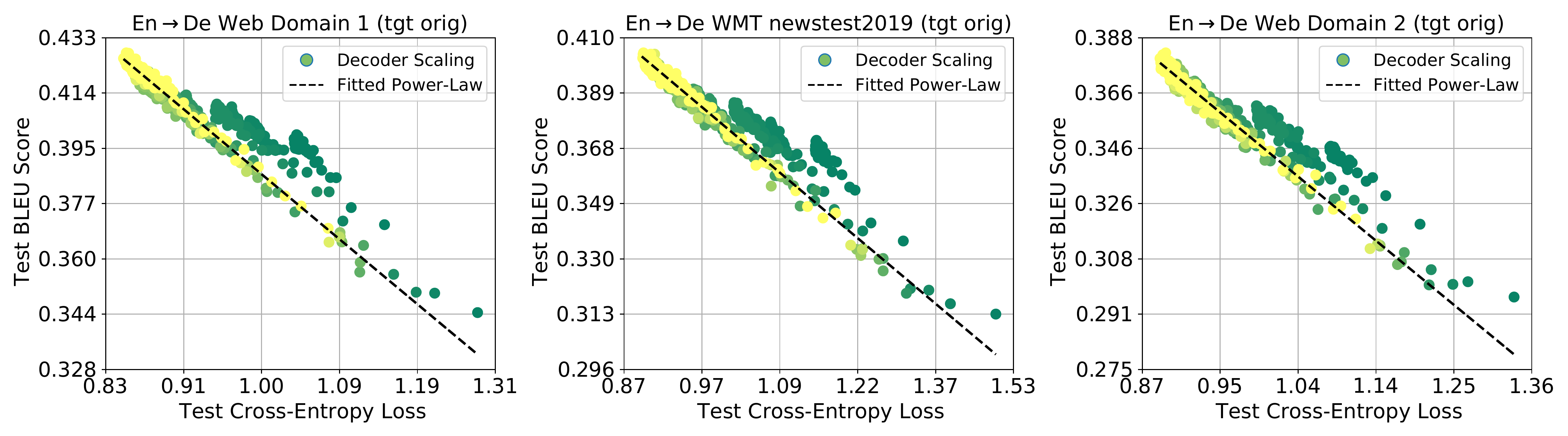}
  \includegraphics[width=\textwidth]{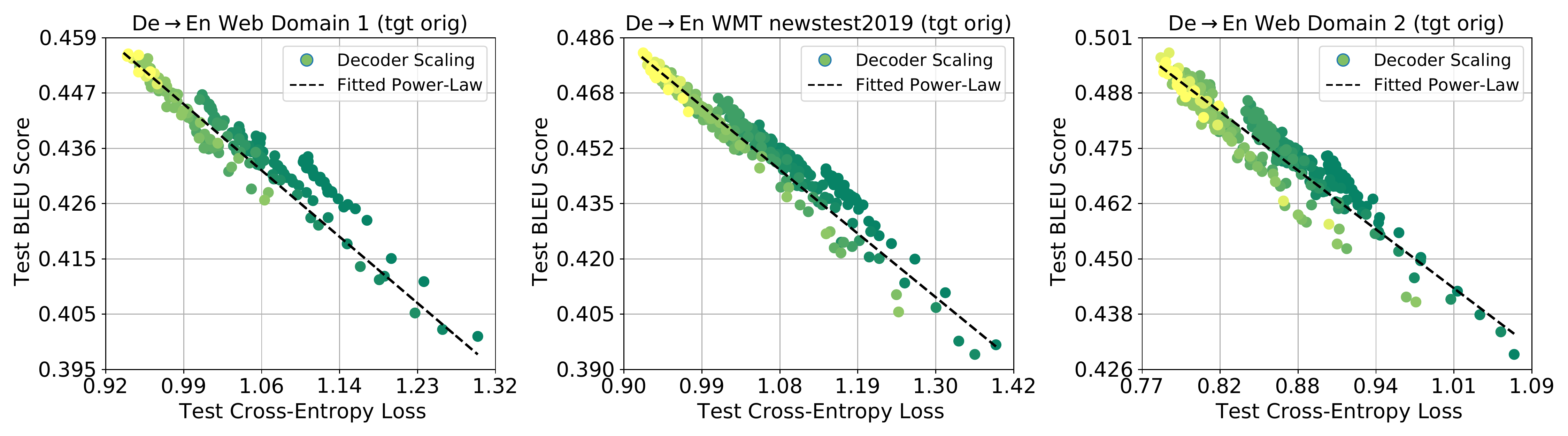}
  \caption{Models with shallow decoders tend to outperform predictions of Eq. \eqref{eq:bleu_loss}. Points with dark green color represent different checkpoints of 6L2L, 6L4L, and 6L6L models. \label{fig:bleu_loss_deviations}}
\end{figure}

\begin{figure}[h]
  \centering
  \includegraphics[width=\textwidth]{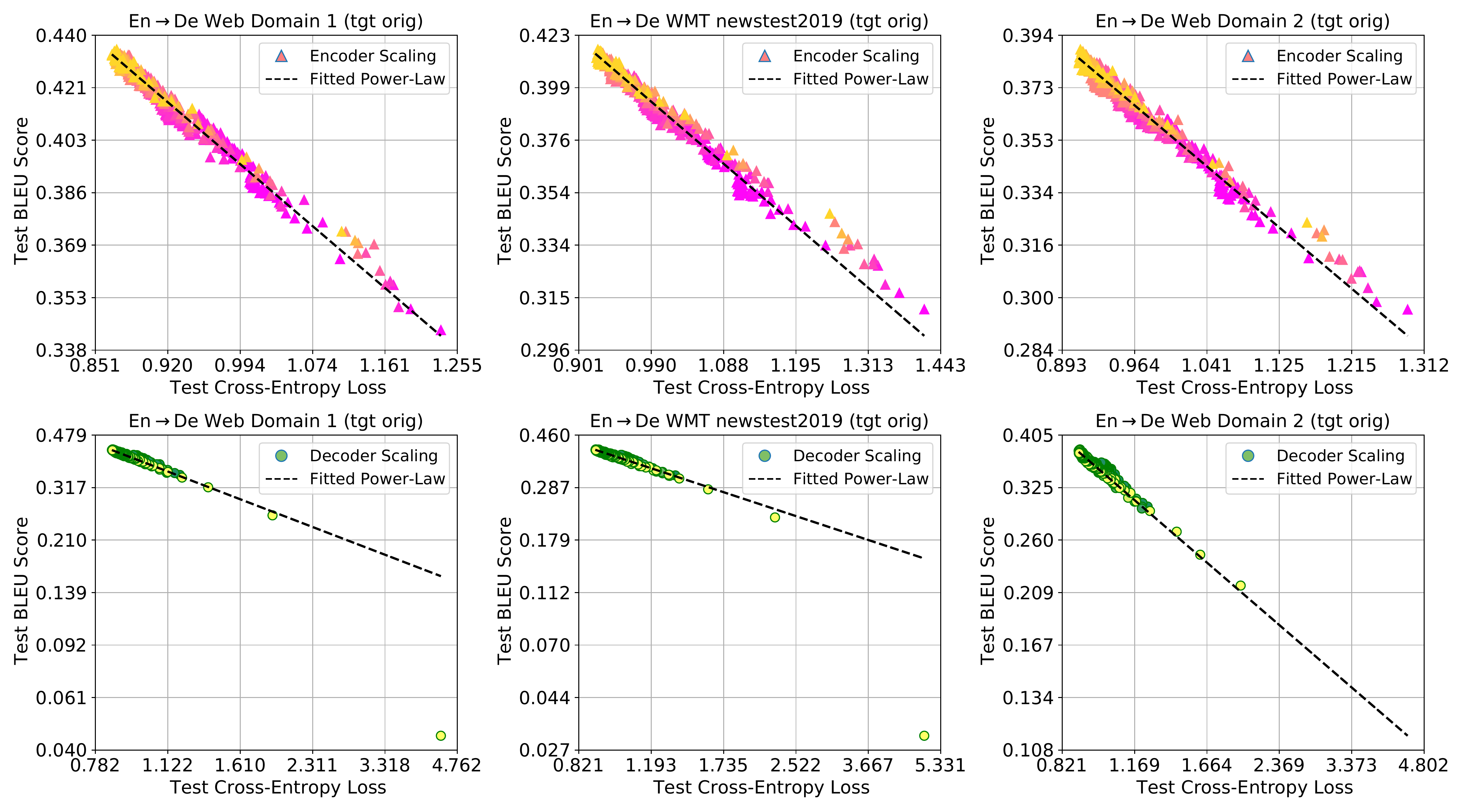}
  \caption{On some of the test sets, data points corresponding to early training checkpoints exhibit deviations from the overall trend. \label{fig:bleu_loss_deviations_2}}
\end{figure}

Finally, we observe a number of deviations from the predictions of Eq. \eqref{eq:bleu_loss}. In particular, models with shallow decoders (6L2L, 6L4L, 6L6L) seem to outperform the trend (Figure \ref{fig:bleu_loss_deviations}). Moreover, we observe that models in the beginning of the training process tend to deviate from the overall trend (Figure \ref{fig:bleu_loss_deviations_2}). We postpone an in-depth analysis of these phenomena to future work. 

\begin{figure}[h]
  \centering
  \includegraphics[width=\textwidth]{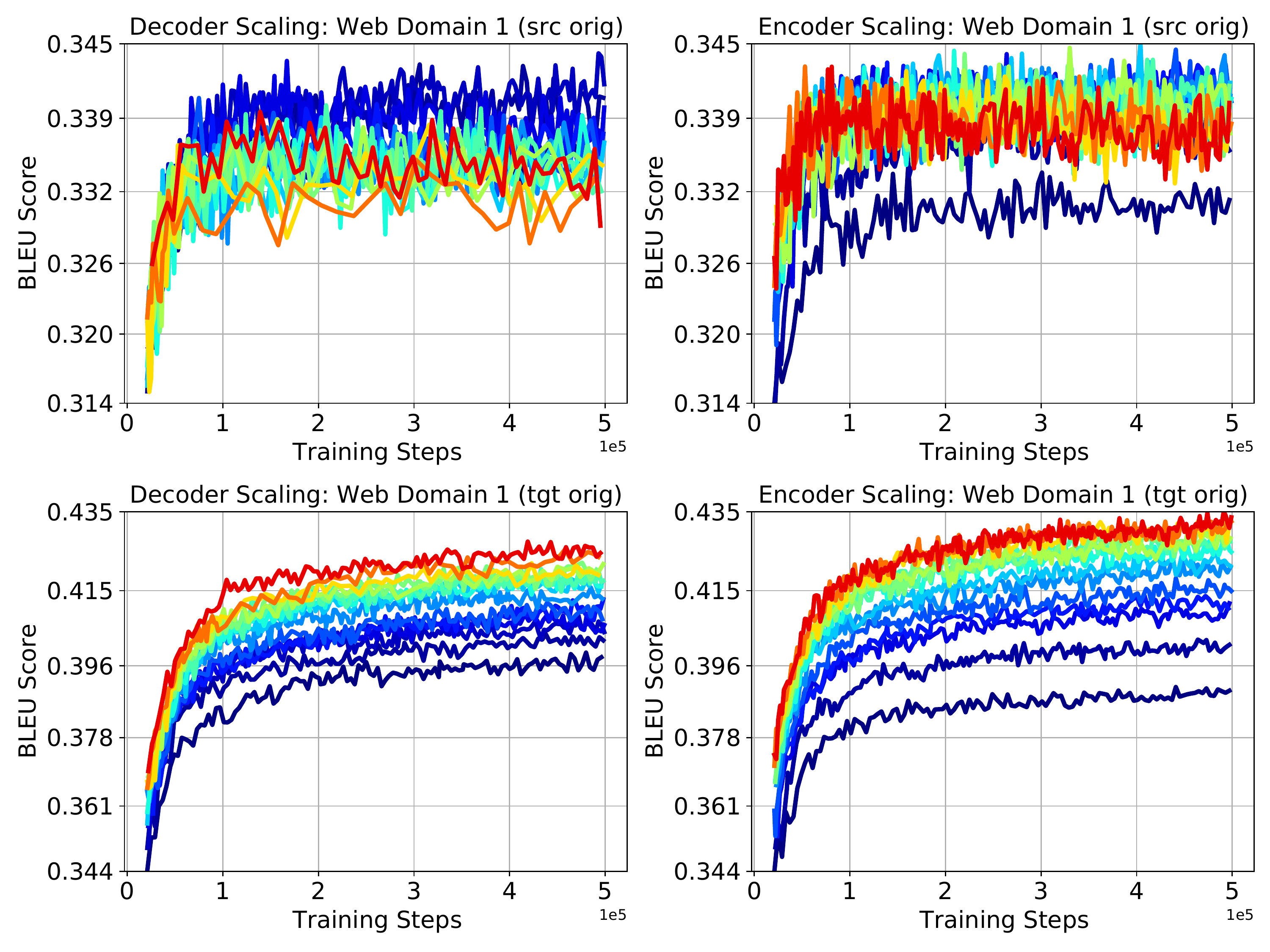}
  \caption{The evolution of BLEU score during the training for English-to-German Web Domain test sets. Warmer colors correspond to larger models. Top row: On source original test data, our largest models achieve lower BLEU scores compared to mid-sized models throughout the training. Bottom row: On target original test data, increasing the model size yields consistent improvements in BLEU score throughout the training. \label{fig:bleu_loss_EnDe}}
\end{figure}

\begin{figure}[h]
  \centering
  \includegraphics[width=\textwidth]{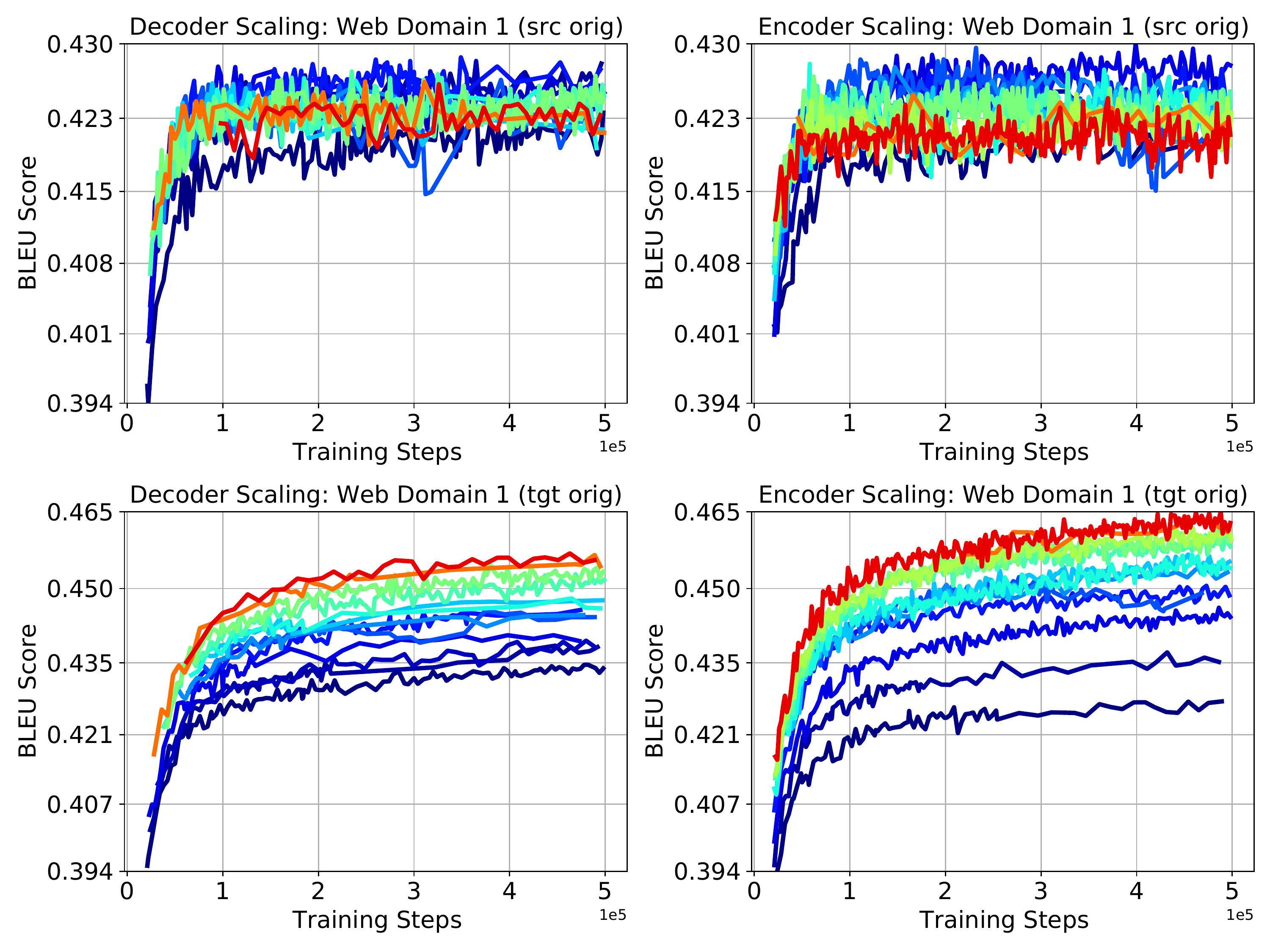}
  \caption{The evolution of BLEU score during the training for German-to-English Web Domain test sets. Warmer colors correspond to larger models. Top row: On source original test data, our largest models achieve lower BLEU scores compared to mid-sized models throughout the training. Bottom row: On target original test data, increasing the model size yields consistent improvements in BLEU score throughout the training. \label{fig:bleu_loss_DeEn}}
\end{figure}

\begin{figure}[h]
  \centering
  \includegraphics[width=\textwidth]{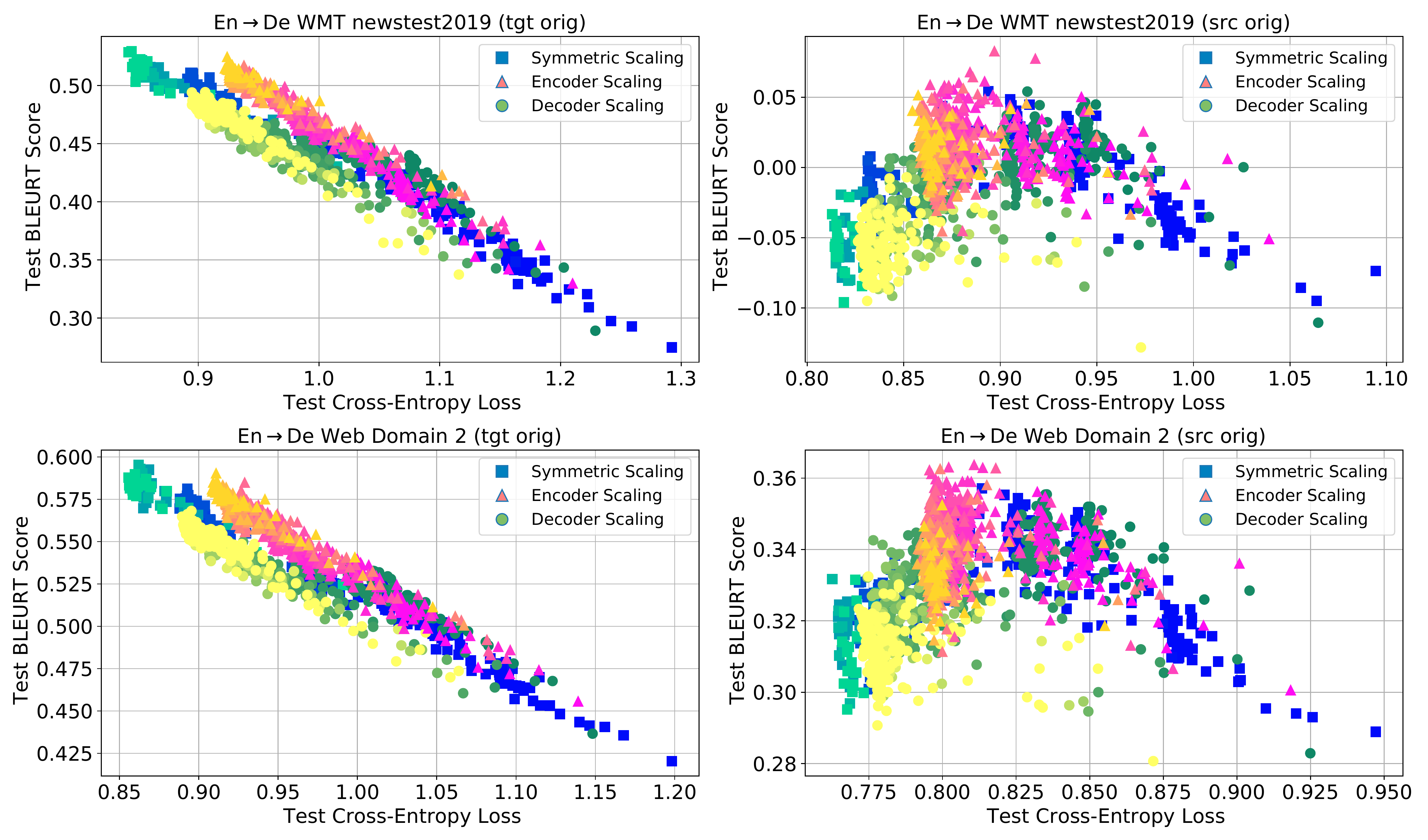}
  \includegraphics[width=\textwidth]{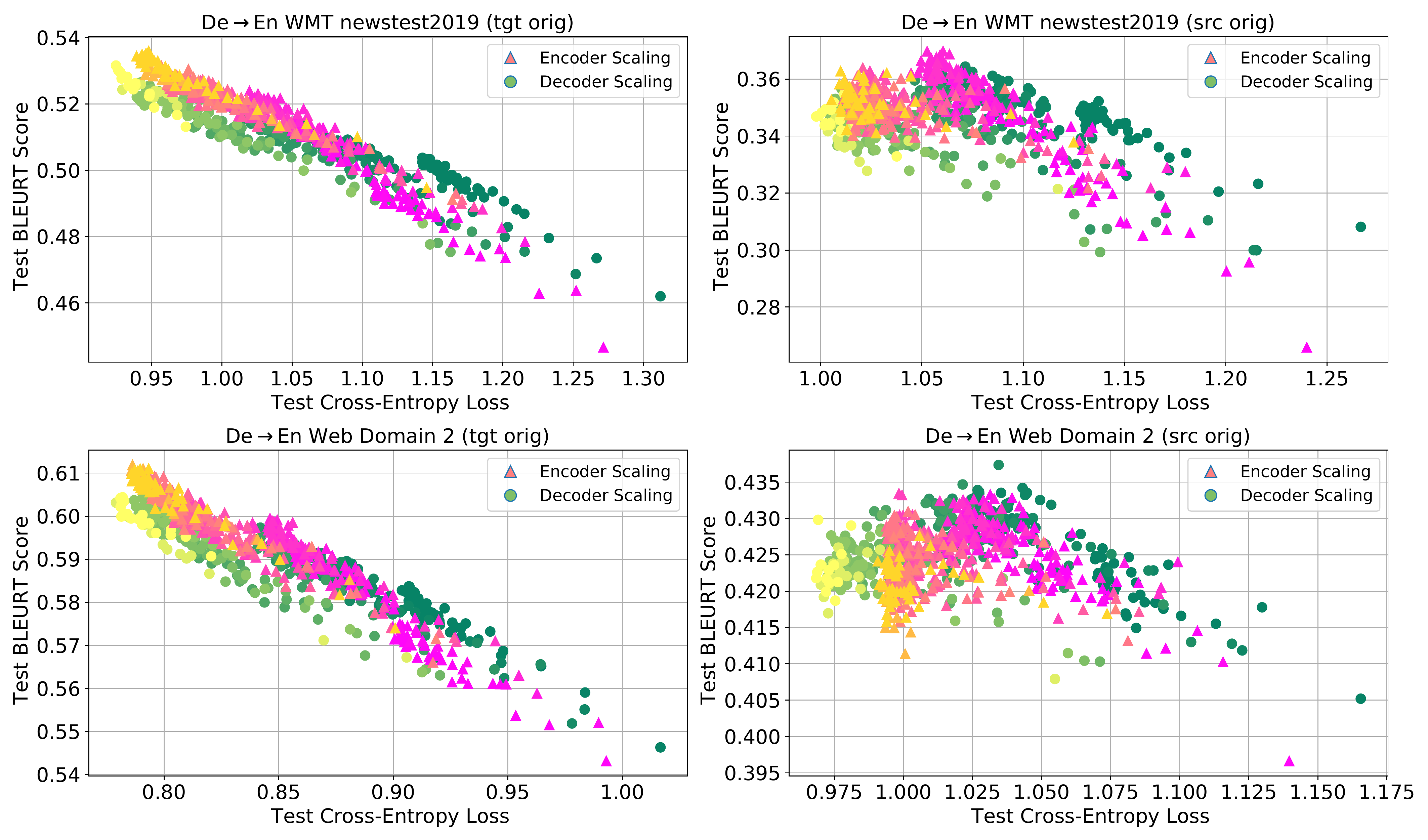}
  \caption{The evolution of BLEURT score as a function of cross-entropy loss for different models. For each scaling approach, warmer colors represent larger models. Each individual color represents different checkpoints of a single model during training. On target original data (left column), improvements to cross-entropy loss lead to consistent improvements in BLEURT score. This relationship breaks down for source original data (right column). \label{eq:bleurt_loss_all}}
\end{figure}

\end{document}